\documentclass[10pt,twocolumn,letterpaper]{article}

\usepackage{cvpr}              
\usepackage{multirow}
\usepackage{pifont}
\usepackage{makecell}
\usepackage{xcolor}
\usepackage[table]{xcolor}
\usepackage{amssymb}
\usepackage{pifont}
\usepackage{tabularx}
\newcommand{\tabincell}[2]{\begin{tabular}{@{}#1@{}}#2\end{tabular}}
\newcommand{\cmark}{\ding{51}}%
\newcommand{\xmark}{\ding{55}}%
\usepackage{float}

\definecolor{darkgreen}{RGB}{0,150,0}

\definecolor{cvprblue}{rgb}{0.21,0.49,0.74}
\usepackage[pagebackref,breaklinks,colorlinks,allcolors=cvprblue]{hyperref}

\def\paperID{4174} 
\def\confName{CVPR}
\def\confYear{2026}

\title{Towards Visual Query Localization in the 3D World}

\renewcommand{\thefootnote}{\arabic{footnote}}
\renewcommand{\thefootnote}{\fnsymbol{footnote}} 

\author{
Liang Peng$^{1}$\footnotemark[1],
Bohan Tan$^{2}$\footnotemark[1],
Zhipeng Zhang$^{2,3}$\footnotemark[3],
Haobo Li$^{2}$,
Yifan Jiao$^{4,5}$,
Xingping Dong$^{1}$\footnotemark[2],
Libo Zhang$^{4,5}$\\
$^{1}$Wuhan University \quad
$^{2}$AutoLab, SAI, Shanghai Jiao Tong University \quad
$^{3}$Anyverse Dynamics\\
$^{4}$University of Chinese Academy of Sciences \quad
$^{5}$Institute of Software, Chinese Academy of Sciences\\[-2pt]
\parbox{\linewidth}{\centering\ttfamily\footnotesize
2025102110090@whu.edu.cn, bohant@hust.edu.cn
}
}

\begin{document}

\twocolumn[{%
\renewcommand\twocolumn[1][]{#1}%
\maketitle
\begin{center}
    \centering
    \captionsetup{type=figure}
    \includegraphics[width=0.99\textwidth]{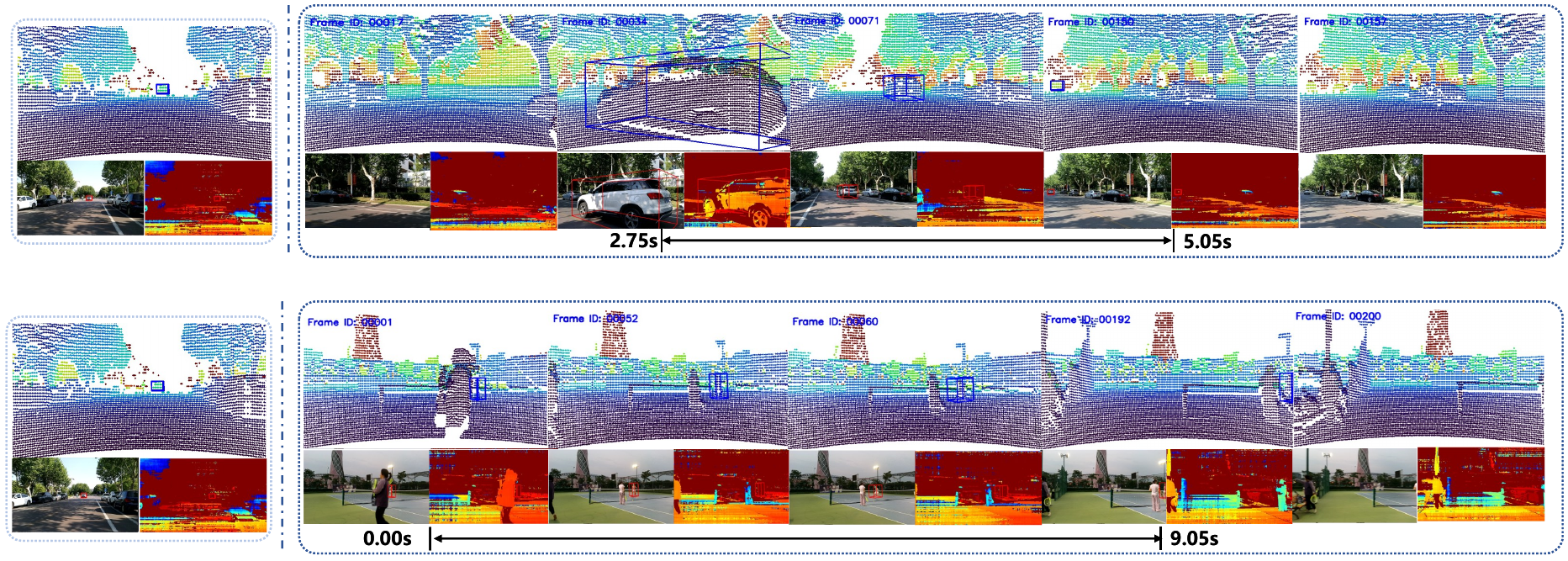}   
    \captionof{figure}{Illustration of 3DVQL that extends visual query localization from 2D videos to 3D multimodal spaces. Each example sequence includes \emph{point cloud}, \emph{RGB image}, and \emph{depth} for 3D visual query localization. \emph{Best viewed in pdf for all figures throughout the paper. }}
    \label{fig:exa}
\end{center}%
}]

\maketitle
\footnotetext[1]{Equal contribution. \quad $\dagger$ First corresponding author; Supported in part by the New Generation Artificial Intelligence National Science and Technology Major Project (No.~2025ZD0123501) and the National Natural Science Foundation of China (Grant No.~62471342, No. 62503323). \\ $\ddagger$ Second corresponding author.}

\renewcommand{\thefootnote}{\arabic{footnote}}
\setcounter{footnote}{0}


\renewcommand{\thefootnote}{\arabic{footnote}}
\setcounter{footnote}{0}

\renewcommand{\thefootnote}{\arabic{footnote}}
\setcounter{footnote}{0}

\begin{abstract}
Visual query localization (VQL) aims to predict a spatial-temporal response of the most recent occurrence from a sequence given a query. Currently, most research focuses on visual query localization from 2D videos, while its counterpart in 3D space has received little attention. In this paper, we make the first attempt to visual query localization in the 3D world by introducing a novel benchmark, dubbed 3DVQL. Specifically, 3DVQL contains 2,002 sequences with around 170,000 frames and 6.4K response track segments from 38 object categories. Each sequence in 3DVQL is provided with multiple modalities including point clouds (PC), RGB and depth images to support flexible research. To ensure high-quality annotation, each sequence is manually annotated with multiple rounds of verification and refinement. To our best knowledge, 3DVQL is the first benchmark towards 3D multimodal visual query localization. To facilitate comparison for subsequent research, we implement a series of representative 3D multimodal VQL baselines using PC and RGB. The experimental results show that existing methods exhibit significant performance variations across different fusion modules. To encourage future research, we propose a lift and attention fusion algorithm named LaF, which significantly outperforms existing baseline models. Our benchmark and model will be publicly released at our webpage \href{https://github.com/wuhengliangliang/3DVQL}{https://github.com/wuhengliangliang/3DVQL}.
\end{abstract}

\section{Introduction}
\label{sec:intro}

Visual Query Localization or VQL is a fundamental challenge in video understanding and embodied intelligence. This task involves locating an object in a lengthy, first-person video using only a single image~\cite{li2024mvbench, ma2025position, li2023uniformerv2, wang2024embodiedscan}. Fueled by large-scale datasets like Ego4D~\cite{grauman2022ego4d}, recent 2DVQL frameworks~\cite{xu2022negative,xu2023my,jiang2023single,fan2025prvql} have achieved excellent localization accuracy. However, their performance is inherently capped by a reliance on the 2D plane, failing to resolve ambiguities from real-world complexities such as target occlusions, drastic appearance shifts, and changing viewpoints, creating a disconnect with our inherently 3D world. We argue that leveraging 3D spatial information like point clouds to redesign VQL is a necessary step forward. This vital research direction remains largely untapped, primarily because of the absence of a suitable benchmark. Therefore, creating a multimodal 3D visual query localization dataset is a critical and urgent necessity to bridge this gap.


\vspace{0.5em}
\noindent
\textbf{Benchmark and Dataset Contributions.} Our primary contribution is 3DVQL, a new benchmark designed to pioneer research in multimodal 3D visual query localization. It contains 2,002 sequences and 6.4K track segments across 18 diverse environments, covering 38 object classes. In terms of scale, as shown in Tab.~\ref{tab:benchmark}, 3DVQL is on par with leading 2D benchmarks in sequence count yet offers nearly double the number of track segments. This achievement is particularly noteworthy given the significantly higher difficulty and cost associated with 3D data collection and annotation. Crucially, each sequence features aligned multimodal data including \emph{point clouds (PC)}, \emph{RGB images}, and \emph{depth} from heterogeneous sensors (Fig.~\ref{fig:exa}), establishing a unified platform to evaluate everything from PC-only localization to multimodal approaches like RGB-PC and RGB-D. We ensured annotation quality through a rigorous process where all sequences were manually labeled with framewise 9 DoF 3D boxes and subjected to multiple rounds of inspection. To our knowledge, 3DVQL is the first benchmark of its kind, uniquely built for 3D visual query localization and the only one supporting both single-modal and multimodal evaluation.

\noindent \textbf{Baseline Construction Contributions.} To characterize the challenges of 3DVQL, we established a suite of baseline models adapted from the 2D VQLoC~\cite{jiang2023single} architecture, with a specific focus in this paper on a dual-modal PC-RGB fusion scheme. Our analysis reveals a fundamental divergence from 2DVQL. For instance, while state-of-the-art 2D methods rely on deeper networks to enhance feature discriminability~\cite{fan2025prvql,jiang2023single}, this strategy proves unstable and can even degrade performance on 3DVQL. This suggests the primary bottleneck is not the discriminability of single-modal features, but rather the challenges of cross-modal observability and spatiotemporal alignment. Neither modality is sufficient on its own for stable, long-horizon localization, as RGB imagery is susceptible to motion blur and occlusion, while point clouds suffer from long-range sparsity, non-rigid deformation, and a lack of fine-grained appearance details. Consequently, effective fusion becomes the critical factor. Success requires careful design, including the explicit use of geometric constraints, the modeling of spatiotemporal consistency, and a deliberate strategy for feature alignment and complementarity. The significant performance variations we observed across different fusion methods underscore our key finding that designing a robust and generalizable fusion module is a more critical challenge than simply optimizing single-modal backbones.

A central challenge in 3DVQL is the effective fusion of 2D image and 3D point cloud data. To this end, we introduce a novel fusion method, namely ``Lift and Fusion (LaF)'', which lifts 2D features into the 3D space, akin to modern Bird's eye view (BEV) perception algorithms in autonomous driving. Specifically, it expands each 2D pixel token into a ray along its line of sight in 3D. To ensure relevance and efficiency, cross-modal attention is restricted to only those 3D voxels falling within the camera frustum. The module then performs multi-head attention along this depth axis, precisely aligning the two modalities based on viewing geometry. Following fusion, a Transformer decoder injects the query embedding into the aligned voxel features. To capture object dynamics, a spatiotemporal Transformer encoder with a local time-window mask models dependencies across the sequence, enhancing robustness while managing computational cost. Finally, the features are upsampled to a fixed resolution, where a 3D prediction head directly regresses the target 9 DoF pose of the most recent and a confidence score.


In summary, our contributions are as follows: \ding{171} We propose 3DVQL, a new benchmark with 2,002 sequences and more than 170K annotated frames, aiming to promote research on 3D visual query localization; \ding{170} 3DVQL provides multiple modalities for each sequence, making it a general platform that supports various visual query localization tasks; \ding{168} We design a series of RGB–PC multimodal baseline models based on the VQLoC architecture to support rigorous comparative experiments and to promote subsequent research on 3DVQL. \ding{169} We introduce a more efficient and concise 3D visual query localization algorithm, LaF, and achieve the best performance up to now.

\renewcommand{\arraystretch}{0.95}
\begin{table}[!t]\small
  \centering
  \caption{Comparison between 2DVQL and 3DVQL. PC: Point cloud, D: Depth.}
  \label{tab:benchmark}
  \vspace{-2mm}
  \setlength{\tabcolsep}{5pt}
  \resizebox{\columnwidth}{!}{%
  \begin{tabular}{*{9}{c}}  
    \Xhline{1.2pt}
    \multirow{2.5}{*}{Benchmark} &
    \multirow{2.5}{*}{\makecell{Tot.\\Seq.}} &
    \multirow{2.5}{*}{\makecell{Ann.\\fr.}} &
    \multirow{2.5}{*}{\makecell{Avg\\trk.}} &
    \multirow{2.5}{*}{\makecell{Res \\trk.}} &
    \multirow{2.5}{*}{\makecell{Obj.\\cls.}} &
    \multicolumn{3}{c}{Mod.} \\
    \cmidrule(lr){7-9}
     &  &  &  &  &  & RGB & PC & D \\
    \hline\hline
    2DVQL~\cite{grauman2022ego4d} & 2{,}538 & --   & 90 & 3.2k & --  & \cmark & \xmark & \xmark \\
    \rowcolor[HTML]{DDDDDD}
    \textbf{3DVQL} & \textbf{2{,}002} & \textbf{170k} & \textbf{84} &
    \textbf{6.4k} & \textbf{38} & \textbf{\cmark} & \textbf{\cmark} & \textbf{\cmark} \\
    \Xhline{1.2pt}
  \end{tabular}%
  }
  \vspace{-3mm}
\end{table}

\section{Related Work}
\label{sec:related}

\textbf{2D Visual Query Localization.} In recent years, the Ego4D episodic-memory benchmark~\cite{grauman2022ego4d} has attracted wide attention, with 2D Visual Query Localization (2DVQL) emerging as a key task. Given a query image taken from outside the target video, the goal is to locate the most recent spatiotemporal occurrence of the queried object in long egocentric videos and return the response track.

Mainstream solutions~\cite{xu2023my,xu2022negative,jiang2023single,khosla2025relocate} fall into multi-stage and single-stage end-to-end paradigms. The multistage pipeline first performs framewise detection and matches candidates to the query, then searches for the latest peak along the temporal axis, and finally applies bidirectional tracking to complete the trajectory. This route benefits from mature detection and tracking modules and is straightforward to engineer, but generating and comparing per-frame proposals is computationally expensive, and occlusion, blur, or abrupt scale changes can cause track fragmentation and error accumulation. To mitigate these issues, follow-up work such as~\cite{xu2022negative} introduces negative samples during training to suppress false peaks over time, while set-based Transformers~\cite{xu2023my} jointly model multiple candidates within the same frame to reduce confusion among similar objects and improve frame-level accuracy. Despite these improvements, maintaining many candidates increases memory and compute costs, and robustness to long missing segments remains limited.

The single-stage end-to-end paradigm targets efficiency and tighter modeling. VQLoC~\cite{jiang2023single} first extracts features for the query and the video with DINO~\cite{oquab2023dinov2}, aligns query–frame spatial correspondences via cross-attention, propagates and refines these correspondences within a local temporal window using a spatiotemporal Transformer, and finally uses a unified prediction head to output frame-level presence probabilities together with anchor-based box regression. This design avoids redundant per-frame proposals and heavy post-processing, improving inference efficiency and scalability while preserving accuracy. Building on this line, PRVQL~\cite{fan2025prvql} adopts a progressive hybrid pipeline comprising global retrieval to quickly localize candidate time spans, local refinement within a temporal window with bidirectional feature refinement to complete trajectories, and an online memory to counter drift caused by long gaps and hard negatives. Compared with purely multi-stage or purely single-stage approaches, this pipeline better balances both robustness and efficiency under long-term discontinuities and strong distractors, yielding more stable, reliable detection and localization.

\vspace{0.3em}
\noindent
\textbf{3D Object Localization.} 3D object localization is a core problem in computer vision that estimates a queried target’s position and orientation in 3D space. The output is an oriented 3D bounding box parameterized by degrees of freedom (DoF), either 7 DoF (center coordinates $(x,y,z)$, size $(l,w,h)$, yaw) or 9 DoF (adding pitch and roll). This capability underpins applications such as autonomous driving, robotic grasping, and augmented reality. Benchmarks fall into two primary settings. Autonomous-driving datasets such as KITTI~\cite{geiger2012we}, nuScenes~\cite{caesar2020nuscenes}, Track-it-in-3D~\cite{yang2022towards}, and Waymo~\cite{sun2020scalability}, which feature multi-object scenes, large depth variation, and complex traffic. Indoor RGB-D or multi-view datasets such as SUN RGB-D~\cite{song2015sun} and ScanNet~\cite{dai2017scannet}, which focus on small objects, occlusions, and fine-grained poses. 

3D object detection is central to 3D object localization, aiming to accurately estimate an object’s position and pose in 3D space. Early methods relying on hand-crafted geometric features ($e.g.$, voxel-grid clustering) performed poorly in complex scenes. With deep learning, VoxelNet~\cite{zhou2018voxelnet} applied 3D convolutions to voxelized point clouds for end-to-end learning, markedly improving recognition and localization but with high computational cost. PointRCNN~\cite{shi2019pointrcnn} combines point-level feature learning (based on PointNet~\cite{qi2017pointnet}) with a two-stage refinement scheme, achieving strong results on KITTI~\cite{geiger2012we}. To further balance accuracy and efficiency, anchor-free paradigms were introduced, such as regression-style detectors ($e.g.$, Focals ConvNet~\cite{chen2022focal}) that directly regress object centers and sizes, and CenterPoint~\cite{yin2021center}, which predicts a center heatmap together with size and orientation, making keypoint-based localization suitable for real-time use. To reduce LiDAR cost, RGB-D and monocular RGB approaches estimate 3D boxes from a single sensor. Deep3DBox~\cite{mousavian20173d} infers 3D boxes by combining 2D detections with depth under geometric constraints. Mono3D~\cite{chen2016monocular} leverages category priors ($e.g.$, average vehicle size) to estimate pose from a single view. End-to-end frameworks such as MonoPSR~\cite{ku2019monocular} fuse depth estimation with 3D regression, though their accuracy remains limited by indirect depth cues.

To compensate for the limitations of single modalities, research has shifted toward multimodal fusion. PointPainting~\cite{vora2020pointpainting} first performs semantic segmentation on RGB images, then uses camera–LiDAR calibration to project per-pixel class probabilities onto the point cloud. The resulting semantic probability vectors are concatenated with geometric features and fed to a 3D detector, enabling early fusion of appearance and geometry and improving recall for distant and sparse objects. The PV-RCNN–based point–image alignment strategy~\cite{shi2019point} extracts high-level semantics from images and projects them onto points or BEV grids, which are concatenated with LiDAR features for proposal generation and RoI feature aggregation. BEVFusion integrates multimodal features within a unified BEV representation using spatial attention, which reduces cross-modal misalignment. These approaches achieve strong results on datasets such as nuScenes~\cite{caesar2020nuscenes}.

Existing 3D detection paradigms mainly perform per-frame full scene detection initialized from the first frame, and do not model or evaluate a query driven setting that localizes only the most recent 3D occurrence of a query in long videos. To close this gap, we introduce 3DVQL, which evaluates three objectives over long horizons, presence verification, temporal localization of the most recent occurrence, and precise 9 DoF 3D localization.

\begin{figure*}[!t]
\centering
\includegraphics[width=\linewidth]{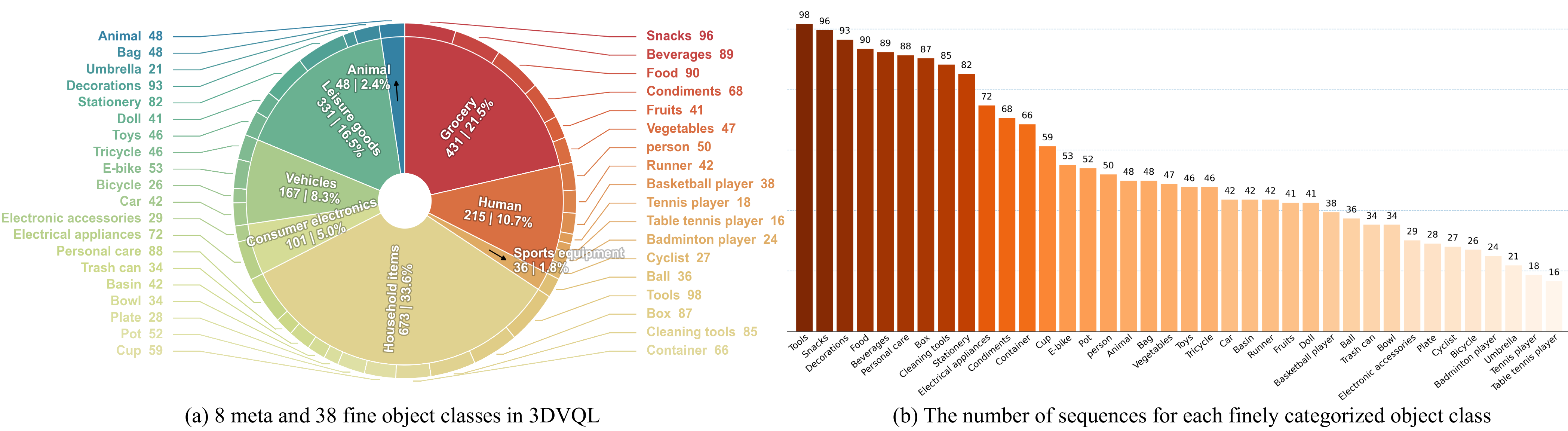}
\caption{\textbf{Meta \& fine-grained categories in 3DVQL.} Figure (a) lists 8 meta categories; Figure (b) shows the ranking of all 38 fine-grained categories by sequence count, grouped by their corresponding meta categories.}
\label{fig:class_statistic}
\vspace{-4mm}
\end{figure*}
\section{The Proposed 3DVQL Benchmark}

\subsection{Construction Principle} 3DVQL aims to serve as a \emph{versatile} platform that advances diverse 3D visual query localization (3DVQL) tasks by providing ample sequences, diverse scenes, rich categories, and high-quality annotations. To this end, we follow the fundamental principles in constructing 3DVQL:
\begin{itemize}
\item \emph{Rich Target Categories and Diverse Scenes.} To build more robust 3D visual query localization, both the training and evaluation phases should include a rich and diverse set of object categories. Accordingly, we expect the benchmark to contain at least 8 categories and 38 subcategories, including common everyday targets suitable for 3D visual query localization.

\item \emph{Different 3D Visual Query Localization Tasks.} To broaden research directions in 3DVQL, sequences should provide multi-modal data, enabling researchers to flexibly choose different input types (single or multiple modalities) to explore various 3D visual query localization tasks according to their specific needs.
\item \emph{Appropriate Scale.} To effectively train and evaluate 3D visual query localization (3DVQL) models, a benchmark must contain a sufficient number of sequences. Given the challenges of collecting and annotating 3DVQL data, we plan to assemble a new benchmark with at least 2k sequences and more than 171k frames. 
\item \emph{Precise Annotation.} Accurate annotation is critical to data fidelity. We manually annotate every point-cloud frame in 3DVQL with higher-precision 9 DoF 3D bounding boxes and perform multiple rounds of meticulous verification and iterative refinement to ensure high-quality labels.
\end{itemize}


\subsection{Data Acquisition.} 

\textbf{Data Acquisition Platform.} We built a mobile robotic system for 3DVQL data collection using a Clearpath Husky A200 base, equipped with a 64-beam LiDAR, a depth camera, and an RGB camera to capture dense geometric and appearance information of the environment. We conducted intrinsic/extrinsic calibration and time synchronization to ensure spatio-temporal consistency across modalities. Operationally, the platform outputs synchronized point clouds and images at 10/20 \emph{fps}. Given 3D visual query localization’s reliance on temporal continuity, we adopt 20 \emph{fps} to increase temporal density and the observability of dynamic processes. Due to space constraints, platform details (including hardware layout and schematics) are provided in the \textbf{supplementary material}.

\noindent
\textbf{Collection of Sequences.} Unlike existing 2DVQL datasets captured with head-mounted (egocentric) cameras~\cite{grauman2022ego4d}, 3DVQL is recorded by a mobile robot across diverse real-world scenes, including streets, parks, university campuses, bedrooms, living rooms, libraries, classrooms, supermarkets, tennis courts, and basketball courts. 3DVQL organizes objects into eight meta categories: \emph{Grocery}, \emph{Person}, \emph{Sports equipment}, \emph{Household items}, \emph{Consumer electronics}, \emph{Vehicles}, \emph{Leisure goods}, and \emph{Animal}. Due to the challenges of 3D data collection and annotation, some categories common in 2DVQL ($e.g.$, static tables and doors) are excluded as they are not suitable for 3DVQL. Within these meta categories, we define 38 fine-grained classes. Fig.~\ref{fig:class_statistic}(a) lists the eight meta categories with their 38 fine-grained classes, and Fig.~\ref{fig:class_statistic}(b) reports the distribution of sequence counts across the fine-grained classes.

To balance geographic and scene diversity, we perform stratified sampling of video clips. To better support the episodic retrieval nature of VQL, we prioritize clips whose queries meet the following criterion. Two successive appearances of the same target are separated by a sufficient temporal interval and accompanied by a substantial spatial or viewpoint change. This typically occurs when the camera wearer or robot briefly passes near the target, leaves for another area, and later returns. In this case, the most recent visible location serves as the retrieval answer.

\vspace{-0.1em}
After defining the categories, we used a mobile robotic platform to collect sequences. To ensure suitability for 3D visual query localization (3DVQL), graduate researchers with 2D VQL expertise assisted acquisition. An expert panel conducted iterative screening of each sequence, removing unsuitable or ambiguous segments. The resulting benchmark comprises 2,002 multimodal sequences (RGB, point clouds, depth), totaling over 170K frames and 6.4K response tracklets. Detailed statistics are provided in Tab.~\ref{tab:benchmark}.

\subsection{Annotation}
To ensure annotation quality of 3DVQL, in terms of query construction, we select a frame from a template video that is independent of the search sequence and exhibits significant appearance differences. Each target corresponds to one query instance. We perform frame wise manual annotation on each query frame and video frame. For each query and its associated video, we annotate the target’s visibility over the entire timeline. When the target is visible, we assign the tightest 9 DoF 3D bounding box that encloses any visible part. When the target is not visible, we mark that period as absent and skip it. Annotations are restricted to contiguous visible segments. Once the target reappears, we continue with the next contiguous segment. For each query video pair, we also record the index of the most recent frame.

In terms of quality control, we assemble a team consisting of domain experts and trained annotators and adopt a multi-stage procedure. First, experts provide demonstration annotations for the query frame and for the first frame of each contiguous segment in the search sequence, and then the annotators complete the frame-wise annotations. Second, experts verify the completed annotations. If consensus is not reached, the sequence is returned to the original annotator for revision. The verification and revision steps are repeated for multiple rounds until acceptance. During verification, we also check temporal consistency and geometric plausibility ($e.g.$, continuity and size stability of the 3D boxes) to further ensure annotation quality. Examples are shown in Fig.~\ref{fig:exa}. More details on the annotation tool, reliability analysis, and extended statistics are provided in the \textbf{supplementary material}.

\subsection{Statistical Characterization of Response Tracks}
\begin{figure}[t]
\centering
\includegraphics[width=\columnwidth]{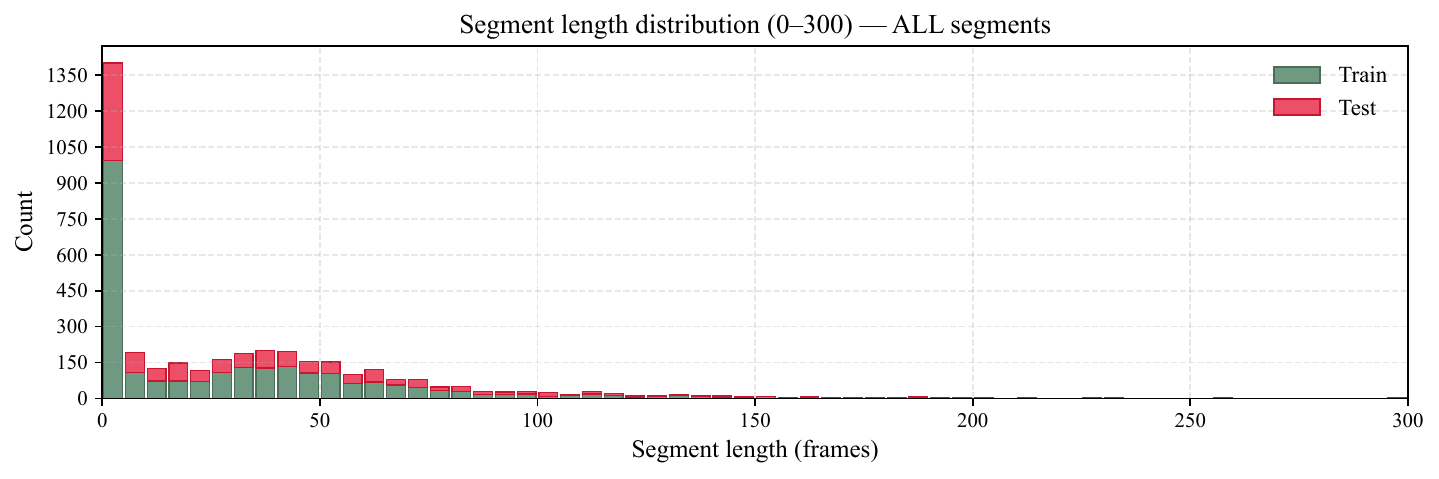}
\caption{Distribution of Query–Response Temporal Distances.}
\label{fig:tracklet}
\vspace{-3mm}
\end{figure}

We compute the \emph{query response separation distance shown in Fig.~\ref{fig:tracklet}}, denoted as
\( d_{\text{sep}} = t_{\text{resp\_end}} - t_{\text{query}} \), to measure how far an algorithm needs to look back along the timeline. The overall distribution is dense at the beginning and sparse later with a clear long tail. In the training set, most samples lie between 0 and 100 frames, while in the test set they concentrate between 0 and 50 frames, as the distance increases, the frequency monotonically decreases and forms a long tail, covering roughly 0 to 250 frames. This indicates that the “most recent occurrence” of most targets can be reached early in the sequence, yet a non-negligible portion appears much later. Therefore, a method should achieve fast hits within short windows and remain stable when performing long-range backtracking and re-detection, maintaining robustness under long-tail cases.

\subsection{Dataset Split, Evaluation Protocol, and Tasks}

\begin{table}[!t]\small
  \centering
  \caption{ Comparison of training and test sets of 3DVQL.}\vspace{-2mm}
    \begin{tabular}{ccccc}
    \Xhline{1.2pt}
          & \tabincell{c}{Total\\Sequences} & \tabincell{c}{Total\\Frames} & \tabincell{c}{Num\\Tracklets} & \tabincell{c}{Object\\Classes} \\
    \hline\hline
    3DVQL$_{\text{Tra}}$ & 1601 & 131.4k & 5157 & 38 \\
    3DVQL$_{\text{Tst}}$ & 401  & 39.6k  & 1319 & 38 \\
    \Xhline{1.2pt}
    \end{tabular}%
  \label{tab:split}%
  \vspace{-3mm}
\end{table}

\textbf{Dataset Split.} We partition 3DVQL into training and test sets following an approximately 80/20 split. Specifically, the training set contains 1{,}601 multimodal sequences, denoted as 3DVQL$_{\text{Tra}}$, and the test set contains 401 sequences, denoted as 3DVQL$_{\text{Tst}}$. Both subsets cover all 38 object categories and preserve the scenario distribution of the full dataset (18 scenarios), which helps reduce evaluation bias due to class or scenario imbalance.

For the split strategy, we adopt stratified sampling to align key statistics as closely as possible, including the length distribution of response tracks, distribution of query-to-response separation distances, and the distributions of object size and motion magnitude, so 3DVQL$_{\text{Tra}}$ and 3DVQL$_{\text{Tst}}$ remain comparable along these dimensions. 

Tab.~\ref{tab:split} summarizes core statistics of the two subsets, including the numbers of sequences, track segments, and frame-level annotations, and provides an overview of category and scenario coverage. More detailed split information ($e.g.$, per-category and per-scenario counts and side-by-side distribution visualizations) will be released together with the paper, dataset, and supplementary materials to facilitate reproducibility and fair comparison.

\vspace{0.45em}
\noindent
\textbf{Evaluation Protocol.} 
\noindent Inspired by the 2DVQL evaluation protocol in~\cite{grauman2022ego4d}, we define the following metrics for 3D localization under the Top-1 retrieval setting, and compute AP by ranking predictions by confidence and integrating precision–recall as standard.
\vspace{0.45em}

\textbf{Temporal Average Precision (tAP).} tAP measures how well the temporal extent of the prediction matches the ground-truth response track. Following the ActivityNet-style protocol~\cite{caba2015activitynet}, we compute mean Average Precision (mAP) over a set of temporal IoU (tIoU) thresholds. We evaluate the tAP at four tIoU thresholds $\{0.25, 0.50, 0.75, 0.95\}$, as well as their average value.

\vspace{0.45em}
\textbf{3D Spatio-Temporal Average Precision (3D-stAP).} 3D-stAP measures how well the 3D spatio-temporal extent of the prediction matches the ground truth response track. We first compute per-frame 3D IoU between the predicted and ground-truth \emph{9 DoF} oriented 3D boxes, then aggregate over time to obtain the 3D spatio-temporal IoU, denoted \(\mathrm{stIoU}_{3D}\). Given the increased difficulty of 9 DoF localization in 3D space, we additionally include an stIoU threshold of $0.05$. We evaluate stAP at five stIoU thresholds $\{0.05, 0.25, 0.50, 0.75, 0.95\}$ and report their average.

\vspace{0.45em}
\textbf{Success (Succ).} Succ measures whether there is any effective overlap with ground truth. A query is counted as successful if \(\mathrm{stIoU}_{3D} \ge 0.05\). Success is the percentage of such queries.

\vspace{0.45em}
\textbf{Recovery\% (Rec\%).} Rec\% measures how much of the ground-truth response track is recovered. It is defined as the percentage of frames within the ground-truth interval for which the predicted 3D bounding box achieves \(\mathrm{IoU}_{3D} \ge 0.5\) with the ground truth 3D bounding box. This follows the robustness spirit of the VOT challenge metric~\cite{kristan2020eighth}.

\begin{figure*}[!t]
\centering
\includegraphics[width=0.95\textwidth]{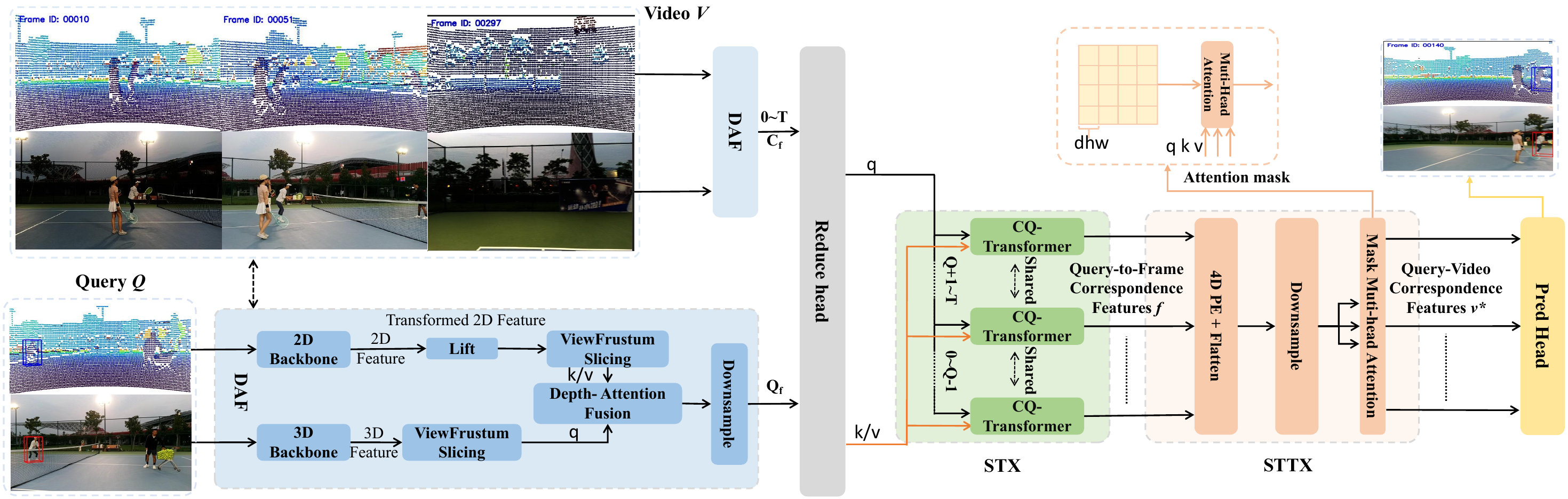}
\caption{Overview of the proposed LaF framework, which employs a DAF module for fusion, followed by STX and STTX modules for spatio-temporal feature extraction, and finally a center-based prediction head for 9 DoF bounding box estimation.}
\label{fig:LaF}
\vspace{-3.0mm}
\end{figure*}
\section{Baselines and Proposed Method}
\subsection{Baselines} To establish performance benchmarks for 3DVQL, we developed a suite of end-to-end RGB-PC baseline models based on the VQLoC~\cite{jiang2023single} architecture. 3D visual query localization is particularly challenging due to the high variability in geometric structures, sparse feature representations, and the inherent complexity of spatio-temporal data. Furthermore, the accurate alignment of multimodal information poses a major challenge. To address these challenges, we propose a foundational pipeline and several variants to investigate different architectural design choices. Unless stated otherwise, all models incorporate the following shared components and default:


\vspace{1.5mm}
\noindent\textbf{Inputs and Encoders.} We use clips of $T$ frames. RGB frames are resized to $448\times448$ and encoded with $C$ output channels ($e.g.$, $C{=}768$ for ViT-B/14~\cite{dosovitskiy2020vit} pretrained with DINOv2~\cite{oquab2023dinov2}). Point clouds are clipped to a fixed workspace $\mathcal{W}$ then voxelized to a dense grid of size $D{\times}H{\times}W = 16^3$ and encoded by 3D sparse convolution module ($e.g.$, $C{=}768$ for 3D backbone pretrained with PV-RCNN~\cite{shi2019pointrcnn}).

\vspace{1.5mm}
\noindent\textbf{Cross-modal Alignment \& Fusion.} For effective feature integration, we first spatially align the 2D features with the 3D voxel grid by lifting them along the depth dimension. The aligned 2D and 3D features are then fused via element-wise addition. This allows the model to incorporate 2D appearance information, while the 3D structure provides the spatial context for alignment.

\vspace{1.5mm}
\noindent\textbf{Query Conditioning.} The query representation is obtained by cropping the fused feature using the 3D query box. This representation then serves as a conditional query to drive a query-conditioned cross-attention over the search features.

\vspace{1.5mm}
\noindent\textbf{Spatial Transformer Module (STX).} This module performs cross-attention, enabling the query feature $Q_f$ to attend to and aggregate relevant spatial cues from $C_f$. The output of this stage is a query-to-frame
correspondence feature, denoted as $f$.

\vspace{1.5mm}
\noindent\textbf{Spatio-Temporal Transformer Module (STTX).} This module performs self-attention, enabling the spatially-enhanced feature $f$ to reason across multiple frames, capturing the temporal consistency and dynamics within the video. This results in a query-video correspondence feature, denoted as $V^*$.

\vspace{1.5mm}
\noindent\textbf{3D Head \& Association.} On a regular 3D grid for decoding (default $16{\times}16{\times}16$), an anchor/center-based head jointly regresses $(\Delta\mathbf{c},\mathbf{s},\mathbf{r})$ (center offsets, size, rotation; 9 DoF) and predicts a presence/consistency score. We select the Top-1 3D box per frame and associate frame-level predictions with the query to form 3D response track segments.

\vspace{1.5mm}
\noindent\textbf{Baseline Models} To explore different design choices, we develop several variants of our baseline by incorporating simplified yet effective modules. \textbf{AnchorFusion-3DVQL (AF):} This variant replaces the center-based detection head with an anchor-based approach and uses a 7 DoF generalized IoU loss instead of the distance loss. \textbf{Guided-Attention Fusion-3DVQL (GAF):} This variant integrates a Guided-Attention Fusion module, which uses depth positional encoding and depth-aware blocks to inject depth cues via cross-attention. \textbf{Projection-Aware Attention Fusion-3DVQL (PAF):} This variant integrates a Projection-Aware Attention Fusion module, where the centers of 3D voxels are projected onto the 2D image plane and corresponding image features are sampled via bilinear interpolation. These sampled 2D features are then fused with the 3D voxel features using a multi-head attention mechanism.

\vspace{1.5mm}
This design follows the query-response trajectory paradigm in VQLoC~\cite{jiang2023single} and extends it to an RGB-PC multimodal setting in 3D space. Variants differ only in the fusion design, temporal window, decoding grid, and head parameterization. We report more detailed ablation experiment analysis in the  supplementary material. 

\vspace{1.5mm}
Notably, our dataset covers everyday scenes in 9 DoF for embodied intelligence applications. Due to the current lack of a 9 DoF IoU operator that supports backpropagation, it is difficult to perform stable IoU supervision over multiple candidate boxes. Therefore, we adopt center-point regression as the training objective. Experimental results demonstrate that the centroid regression-based method can also achieve certain performance progress.

\subsection{Proposed Method} While the baseline and its variants provide a solid framework for multimodal fusion, we find that its feature element-wise addition strategy does not fully exploit the precise geometric correspondence between pixels and points. To this end, we design a depth attention fusion module and integrate it into the VQLoC architecture. We refer to this method as LaF, as shown in Fig.~\ref{fig:LaF}. This method replaces the fusion module in the baseline with a perspective-aware attention  fusion module. Unless otherwise specified, the remaining components and default follow the baseline.

\vspace{-3.0mm}
\paragraph{Lift Module.} The Lift module transforms 2D image features into a 3D volumetric representation. It projects each 2D pixel token along its corresponding viewing ray, generating a set of 3D voxel candidates. To maintain computational tractability, this process is strictly confined to voxels located within the frustum of the camera.

\vspace{-3.0mm}
\paragraph{Depth Attention Fusion Module (DAF).} As displayed in Fig.~\ref{fig:LaF}, the Depth Attention Fusion Module (DAF) employs a perspective-aware multi-head attention mechanism. In this mechanism, the 3D feature serves as queries, while the transformed 2D image features act as keys and values. The attention is computed slice-wise along the depth axis, thereby guaranteeing geometry-aware alignment and enabling adaptive aggregation of 2D information for each 3D query. Subsequently, the resulting fused query feature $Q_f$ is cropped by a 3D region of interest that is precisely aligned with the initial 3D query box.
\vspace{-3.0mm}
\paragraph{Anchor Head and Decoding.} We place a uniform anchor grid $\{\mathbf{a}_n\}_{n=1}^{16^3}$ over $\mathcal{W}$. A CenterPoint-like prediction head takes the upsampled feature volume and, for each frame $t$ and anchor $n$, predicts (i) a center offset, (ii) box size, (iii) rotation and (iv) a presence score. The final 9 DoF prediction of frame $t$ is:
\begin{equation}
n_t^\ast = \arg\max_n p_{t,n}, \quad
\hat{\mathbf{b}}_t = \text{Decode}(\mathbf{a}_{n_t^\ast}, \text{pred}_{t,n_t^\ast}),
\label{eq:decode}
\end{equation}
where \text{Decode} applies the predicted offsets, sizes, angles to the corresponding anchor center.
\vspace{-3.0mm}
\paragraph{Training.}
For frames where the object appears, we mark as positive those anchors whose centers lie within a fixed radius $\tau_c{=}0.3$ m of the ground-truth center and are among the top-$5$ nearest centers. We optimize a regression term on the 9 DoF box parameters together with a focal classification term on the presence score:
\begin{equation}
\mathcal{L} = \lambda_{\text{c}} \mathcal{L}_{c} + \lambda_{\text{s}} \mathcal{L}_{s} + \lambda_{\text{r}} \mathcal{L}_{r} + \lambda_{\text{cls}} \mathcal{L}_{\text{cls}} + \lambda_{\text{dist}} \mathcal{L}_{\text{dist}},
\label{eq:loss}
\end{equation}
where $\mathcal{L}_{c}$, $\mathcal{L}_{s}$ and $\mathcal{L}_{r}$ respectively denote the regression F1 loss for the box parameters center,  size and rotation, $\mathcal{L}_{\text{cls}}$ is the focal loss for classification, and $\mathcal{L}_{\text{dist}}$ penalizes the distance between the ground-truth center and the positive centers after applying the predicted offset.

\begin{figure*}[!t]
  \centering
  \includegraphics[width=0.95\linewidth]{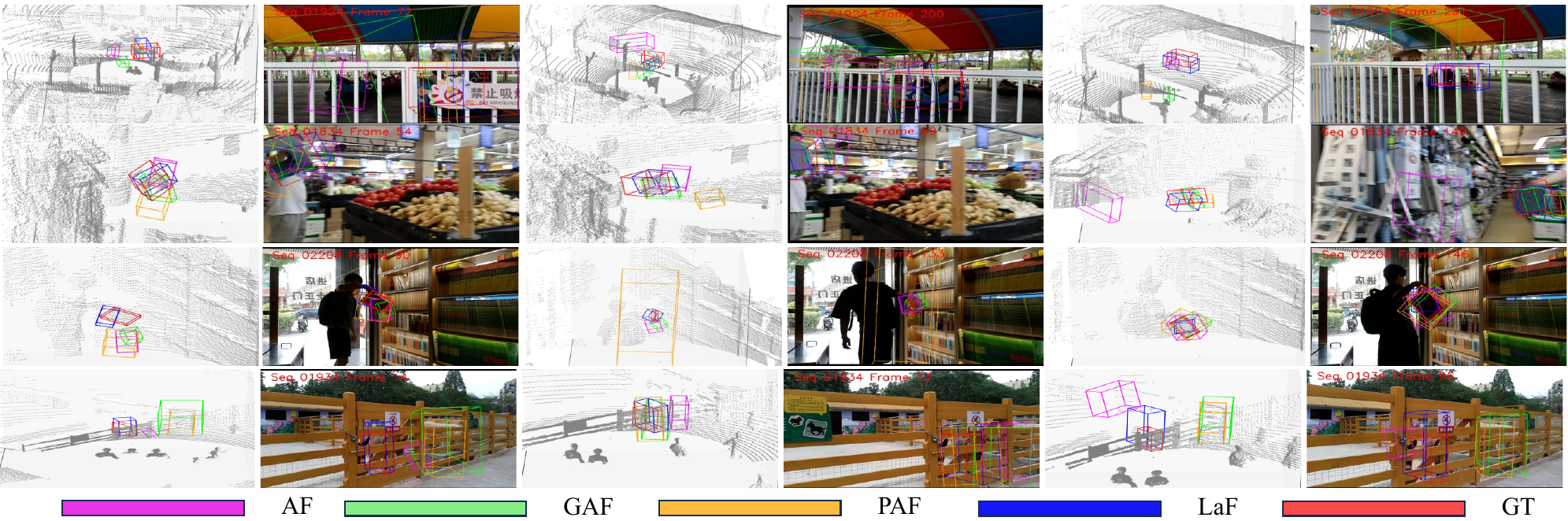} 
  \caption{Qualitative results of our method and comparison to other approaches on four sequences.}
  \label{fig:vis}
  \vspace{-3mm}
\end{figure*}

\section{Experiments}
\paragraph{Implementation.} Our LaF model is implemented in PyTorch with NVIDIA RTX 4090 GPUs. Similarly to VQLoC~\cite{jiang2023single}, we use ViT~\cite{dosovitskiy2020vit} pretrained with DINOv2~\cite{oquab2023dinov2} as the 2D backbone and a 3D sparse convolution module initialized with PV-RCNN’s~\cite{shi2019point} pretrained weights as the 3D backbone. LaF is trained end-to-end for 400 epochs using a peak learning rate of $10^{-4}$ and a weight decay of $5 \times 10^{-2}$. The images are resized to $448 \times 448$, while the point cloud is resampled to 4096 points. The pooling size for RoIAlign is 5. The $\lambda_{c}$, $\lambda_{s}$, $\lambda_{r}$, $\lambda_{cls}$ and $\lambda_{dist}$ are empirically set to 1.0, 1.0, 0.1, 100, 0.3. We sample $T=20$ frames per video clip through random sampling, ensuring a balanced ratio of positive and negative frames. We employ $16 \times 16 \times 16$ centers uniformly distributed within the space range $\mathcal{W} {=} [0,10]\times[-2,2]\times[-1,1]\ \text{m}$. 

\vspace{-3.0mm}
\paragraph{Evaluation Results.}

\begin{table}[!t] 
  \centering
  \scriptsize
  \setlength{\tabcolsep}{1.6mm}
  \renewcommand{\arraystretch}{0.95}
  \caption{Comparison of multiple baselines on the 3DVQL (RGB-PC) test set. The first three rows correspond to our baseline methods, only the last row (our LaF) is highlighted in bold.}
  \vspace{-2mm}
  \resizebox{\linewidth}{!}{
    \begin{tabular}{@{}l c c c c c c@{}}
      \toprule
      Method &
      \multicolumn{1}{c}{tAP} &
      \multicolumn{1}{c}{$\text{tAP}_{0.25}$} &
      \multicolumn{1}{c}{stAP} &
      \multicolumn{1}{c}{$\text{stAP}_{0.05}$} &
      \multicolumn{1}{c}{rec.\%} &
      \multicolumn{1}{c}{Succ.}
      \\
      \midrule
      AF & 0.181 & 0.442 & 0.003 & 0.015 & 0.093 & 11.693 \\
      GAF & 0.291 & 0.597 & 0.015 & 0.075 & 0.049 & 26.309 \\
      PAF & 0.224 & 0.577 & 0.021 & 0.104 & 0.115 & 32.156 \\
      \midrule
      LaF &
      \textbf{{0.293}} &
      \textbf{{0.607}} &
      \textbf{{0.044}} &
      \textbf{{0.222}} &
      \textbf{{0.264}} &
      \textbf{{46.041}} \\
      \bottomrule
    \end{tabular}
  }
  \label{tab:mvql_overall}
  \vspace{-3mm}
\end{table}



To assess the effectiveness of different approaches, we evaluate three baselines (AF, GAF, PAF) and our proposed method LaF on the 3DVQL RGB-PC test set under a unified training and evaluation protocol. Table~\ref{tab:mvql_overall} reports tAP, $\text{tAP}_{0.25}$, stAP, $\text{stAP}_{0.05}$, rec, and Succ; stAP is computed with strict 9 DoF alignment, while $\text{stAP}_{0.05}$ uses a relaxed $0.05$ tolerance. As shown in Tab.~\ref{tab:mvql_overall}, LaF attains the best results on all metrics: tAP $0.293$, $\text{tAP}_{0.25}$ $0.607$, stAP $0.044$, $\text{stAP}_{0.05}$ $0.222$, rec $0.264$, and Succ $46.041$. Relative to the strongest baseline for each metric, the absolute gains are $0.002$ (tAP), $0.010$ ($\text{tAP}_{0.25}$), $0.023$ (stAP), $0.118$ ($\text{stAP}_{0.05}$), $0.149$ (rec), and $13.885$ (Succ).

\begin{table}[!t] 
  \centering
  \scriptsize
  \setlength{\tabcolsep}{1.7mm}
  \renewcommand{\arraystretch}{1.05}
  \caption{Ablation on DAF for LaF on the 3DVQL (RGB-PC) test set. Only the last row (w/ DAF) is highlighted in bold.}
  \vspace{-2mm}
  \resizebox{\linewidth}{!}{
    \begin{tabular}{@{}l c c c c c c@{}}
      \toprule
      Method &
      \multicolumn{1}{c}{tAP} &
      \multicolumn{1}{c}{$\text{tAP}_{0.25}$} &
      \multicolumn{1}{c}{stAP} &
      \multicolumn{1}{c}{$\text{stAP}_{0.05}$} &
      \multicolumn{1}{c}{rec.\%} &
      \multicolumn{1}{c}{Succ.}
      \\
      \midrule
      LaF (w/o DAF) & 0.134 & 0.347 & 0.007 & 0.033 & 0.029 & 18.027 \\
      \midrule
      LaF (w/ DAF) &
      \textbf{{0.293}} &
      \textbf{{0.607}} &
      \textbf{{0.044}} &
      \textbf{{0.222}} &
      \textbf{{0.264}} &
      \textbf{{46.041}} \\
      \bottomrule
    \end{tabular}
  }
  \label{tab:mvql_ablation_fh}
  \vspace{-3mm}
\end{table}
Qualitative comparison on 3DVQL. Fig.~\ref{fig:vis} shows visualization results of all methods in the 9 DoF setting including AF, GAF, PAF, our LaF and GT. For each example we select several frames from the search sequence and draw the 3D bounding boxes in both the point cloud and the RGB image. In most frames the boxes from the other methods do not tightly cover the object and their orientations are clearly off. This shows that these baseline methods are not reliable for 9 DoF spatial localization. In contrast LaF stays closer to the target throughout the sequence and keeps a more reasonable pose, so it is more stable for 9 DoF spatiotemporal localization. Because of space limits more visual results are provided in the supplementary material.


\vspace{-4mm}
\paragraph{Ablation on the Depth Attention Fusion (DAF).}
To isolate the contribution of DAF in LaF, we construct a variant in which DAF is removed. In this variant, the lifted 2D features are directly fused with the 3D features via element-wise addition, and fed into the subsequent pipeline and prediction head, while keeping the Lift module, the remaining network architecture, the training schedule, and the evaluation protocol unchanged (see Tab.~\ref{tab:mvql_ablation_fh}). Comparing this variant with the full LaF shows that introducing DAF consistently improves all metrics, with the largest gain in spatiotemporal localization, where \textbf{stAP} increases from $0.007$ to $0.044$. The full set of results is reported in Tab.~\ref{tab:mvql_ablation_fh}. This further confirms that depth attention fusion is effective in enhancing the performance of 3D query localization.

\section{Conclusion and Limitation} In this paper, we propose 3DVQL for visual query localization in 3D space. It contains 2,002 multimodal sequences and 6.4K response track segments. In addition, to provide baselines for future research, we build a series of multimodal baseline models, establish a unified evaluation protocol, and conduct large-scale evaluations with in-depth analyses. Furthermore, we propose a simple and effective multimodal method for 3D visual query localization, named LaF, and achieve state-of-the-art results. We believe that our benchmark and baselines will promote further research on 3DVQL and its applications. Nevertheless, this work has several limitations. First, the experiments mainly focus on 3DVQL$_{\text{RGB-PC}}$ and do not investigate 3DVQL$_{\text{RGB-D}}$. Second, our sequences are relatively short, which is not suitable for long-term localization. Given that 3DVQL$_{\text{RGB-PC}}$ is the current focus of research and that our primary goal is to provide a new benchmark for 3D visual query localization, we leave richer and long-term 3DVQL to future work.


\vspace{0.3em}
\noindent
\textbf{Acknowledgments.} We would like to thank Shuang Peng, Wei Chen, Ruixian Zhao, and Xiang Zhang for data collection.
\clearpage
\setcounter{page}{1}

\twocolumn[{%
\renewcommand\twocolumn[1][]{#1}%
\maketitlesupplementary
\begin{center}
    \centering
    \captionsetup{type=figure}
    \includegraphics[width=1.0\textwidth]{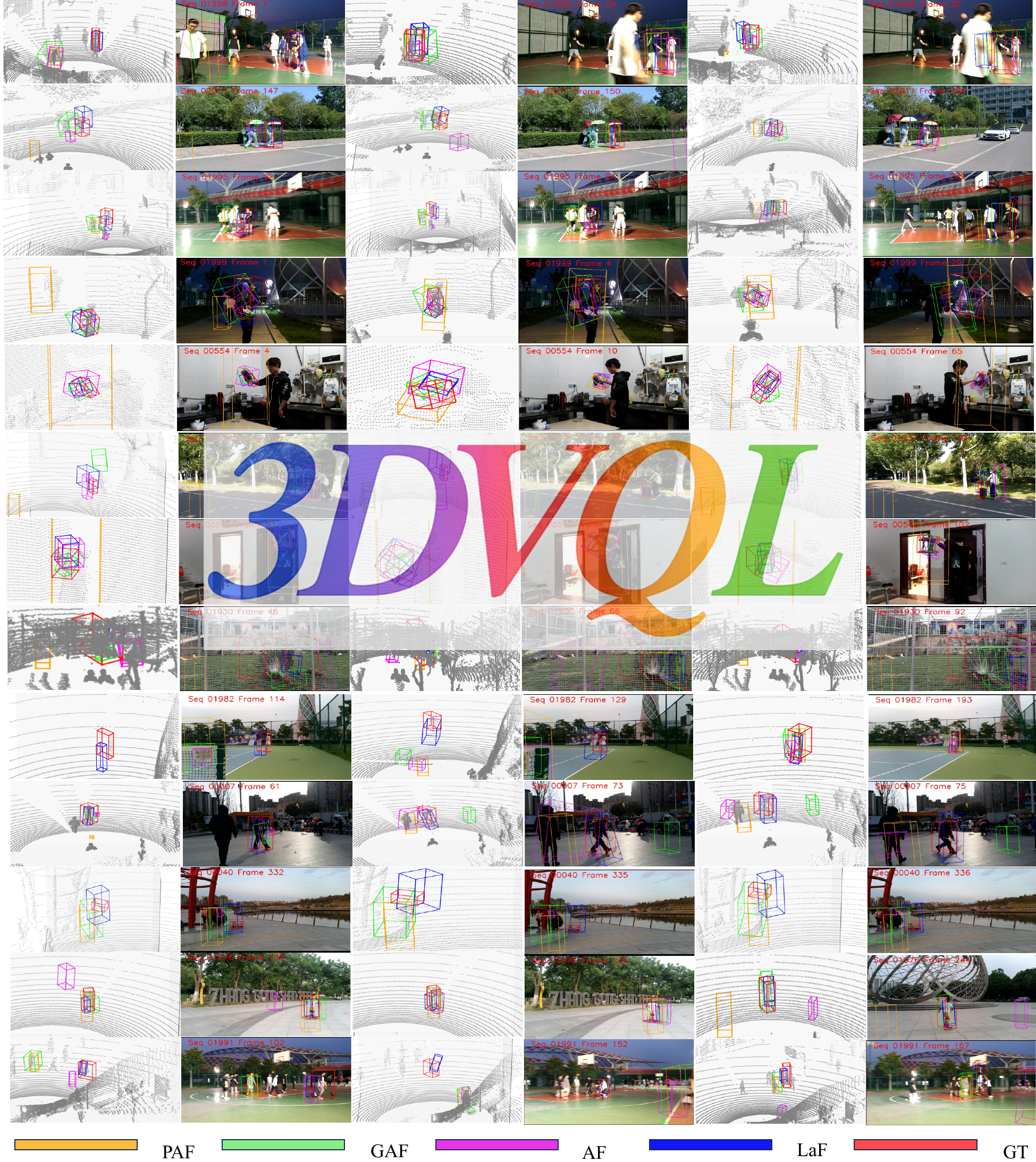}   
    \captionof{figure}{Qualitative results of several baselines and our proposed LaF. We can see that, the proposed LaF locates target object in different scenarios, showing its robustness for 3DVQL.}
    \label{fig:supp_vis}
\end{center}%
}]


\noindent
The following materials provide implementation details, dataset statistics, and additional qualitative visualizations.

\begin{itemize}
	\setlength{\itemsep}{2pt}
	\setlength{\parsep}{2pt}
	\setlength{\parskip}{2pt}
	
	\item[] \textbf{S1 \; Mobile Robotic Platform} \\ In this section, we demonstrate more details of our mobile robotic platform used for multimodal data collection.

	\item[] \textbf{S2 \; Annotation Tool} \\ We display more details of the annotation tool in labeling sequences with 9DoF 3D bounding boxes and its reliability analysis for high-quality annotation.
	

    \item[] \textbf{S3 \; More Statistics} \\ We show the distribution of 3DVQL across spatial locations, as well as more detailed statistical information on the response tracklets. 

     \item[] \textbf{S4 \; Evaluation Metrics and 3D IoU} \\ We describe the formulation of different 3DVQL tasks.

    
    \item[] \textbf{S5 \; Details of Baselines} \\ We present the details of baselines.

    

    \item[] \textbf{S6 \;Qualitative Results} \\ We offer more qualitative analysis of our LaF and its comparison to other trackers on 3DVQL.

    \item[] \textbf{S7 \; Maintenance and Responsible Usage of 3DVQL for Research} \\ We discuss the maintenance and responsible usage of our proposed 3DVQL for research.

\end{itemize}

\section*{S1 \; Mobile Robotic Platform}

\begin{figure}[!t]
    \centering
    \includegraphics[width=0.83\linewidth]{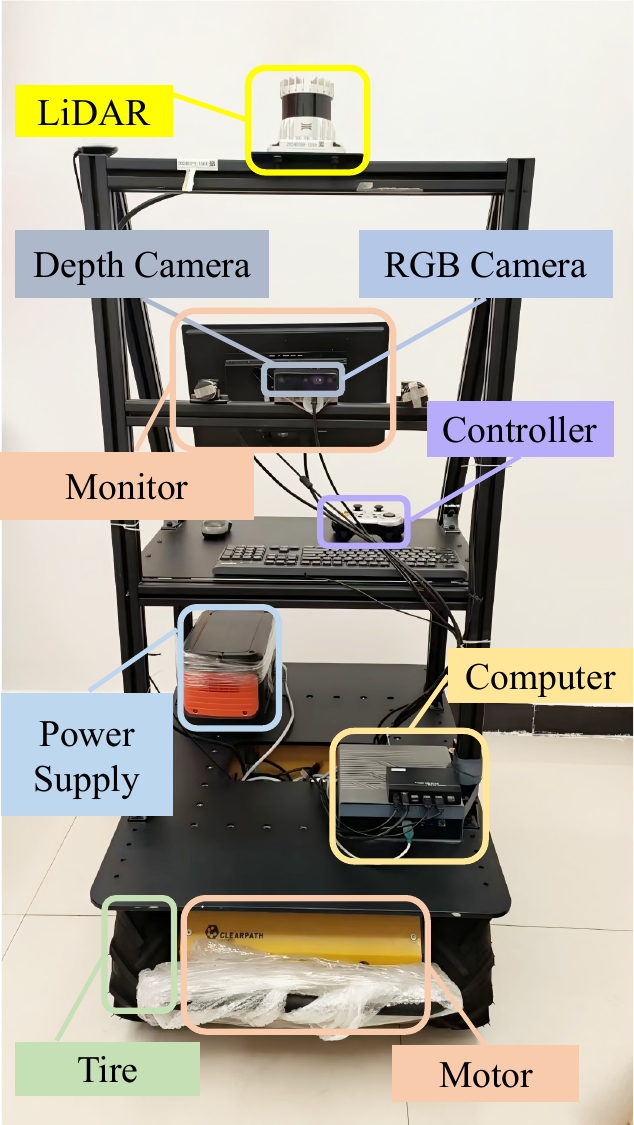}
    \caption{Our mobile robotic platform for data collection.}
    \label{fig:platform}
\end{figure}

To build the multimodal data resources of 3DVQL, we integrated and constructed a mobile robotic platform on the Clearpath Husky A200 chassis. The platform is equipped with a 64-beam LiDAR, an RGB camera, and a depth camera, and with the tool of~\cite{dhall2017lidar} we completed multi-sensor time synchronization and extrinsic calibration to ensure cross-modal geometric consistency. The physical platform is shown in Fig.~\ref{fig:platform}, which displays the mobile robotic platform used for multimodal data acquisition during the development of 3DVQL, and the specific configurations of the sensors and the robot chassis are given in Tab.~\ref{tab:specitic_config}.

\begin{table}[H]\small
    \centering
    \caption{Specific configuration of our mobile robotic platform.}
        \begin{tabular}{ll}
            \Xhline{1.2pt}
            \textbf{Device Name} & \textbf{Specification} \\ 
            \hline
            LiDAR Sensor       & Ouster OS-64 (64-beam)           \\
            Depth Camera       & OAK D-Pro              \\
            RGB Camera         & FLIR BFS-U3-32S4C-C    \\ 
            Robot   Chassis    & Clearpath Husky A200   \\
            \Xhline{1.2pt}
        \end{tabular}
    \label{tab:specitic_config}
\end{table}

\subsection*{S1.1 \; 3DVQL Task Definition and Problem Setting}
In 3DVQL, the input consists of a visual query and a multimodal search sequence, and the goal is to predict both the temporal interval and the frame-wise 9DoF 3D bounding boxes of the queried target. Depending on the sensor configuration, the task can be instantiated as point-cloud-only, RGB-D, RGB-PC, or other multimodal variants. All 9DoF 3D bounding boxes are defined in the camera coordinate system. The RGB camera, depth camera, and LiDAR are geometrically aligned through time synchronization and extrinsic calibration, so boxes and multimodal features can be interpreted in a shared calibrated space.

3DVQL is not equivalent to standard 3D single-object tracking. In 3D SOT\cite{jiao2025gsot3d}, the tracker is typically initialized with the target state in the first frame and then follows a continuously visible target. In contrast, 3DVQL starts from a query template rather than first-frame ground truth, and the target may disappear and reappear across multiple response segments. The model must therefore retrieve and re-localize the queried target under viewpoint changes, distractors, and intermittent visibility, rather than merely propagate a known track.

Compared with Ego4D VQ3D~\cite{grauman2022ego4d}, which mainly focuses on egocentric human videos, 3DVQL targets embodied robot scenarios and provides calibrated RGB, depth, and point cloud data for full 3D localization. Moreover, 3DVQL evaluates stricter 9DoF spatial alignment together with temporal localization, making it better suited for multimodal embodied perception in physical 3D environments.

\subsection*{S1.2 \; Query and Search Sequence Acquisition}
Our query template frame is not sampled from the same video as the search sequence. For each category, we separately collect a template video and a search sequence, where the template sequence is about 80 frames long, and the two are kept independent in data source to reduce the matching bias caused by adjacent temporal information. The template frame used to construct the query is selected from the template video, and the target is required to have complete and clear appearance information. During collection, in order to further ensure effective differences between the template frame and the search sequence, the template video is usually not collected in the same sub-scene as the search sequence, and differences in appearance, motion state, and sub-scene are maintained as much as possible. For example, the target in the search sequence may appear in a forward-walking state, while the template frame may capture its back view. Compared with the target instance in the search sequence, the template frame usually has more obvious differences in viewpoint, position, and surrounding context. This design further increases the difficulty of the task and makes it essentially different from a normal 3DSOT setting, so it can more effectively evaluate the model's query-driven target localization ability and cross-view matching ability in complex scenes.


\section*{S2 \; Annotation Tool}

\begin{figure*}[!t]
    \centering
    \includegraphics[width=1.0\linewidth]{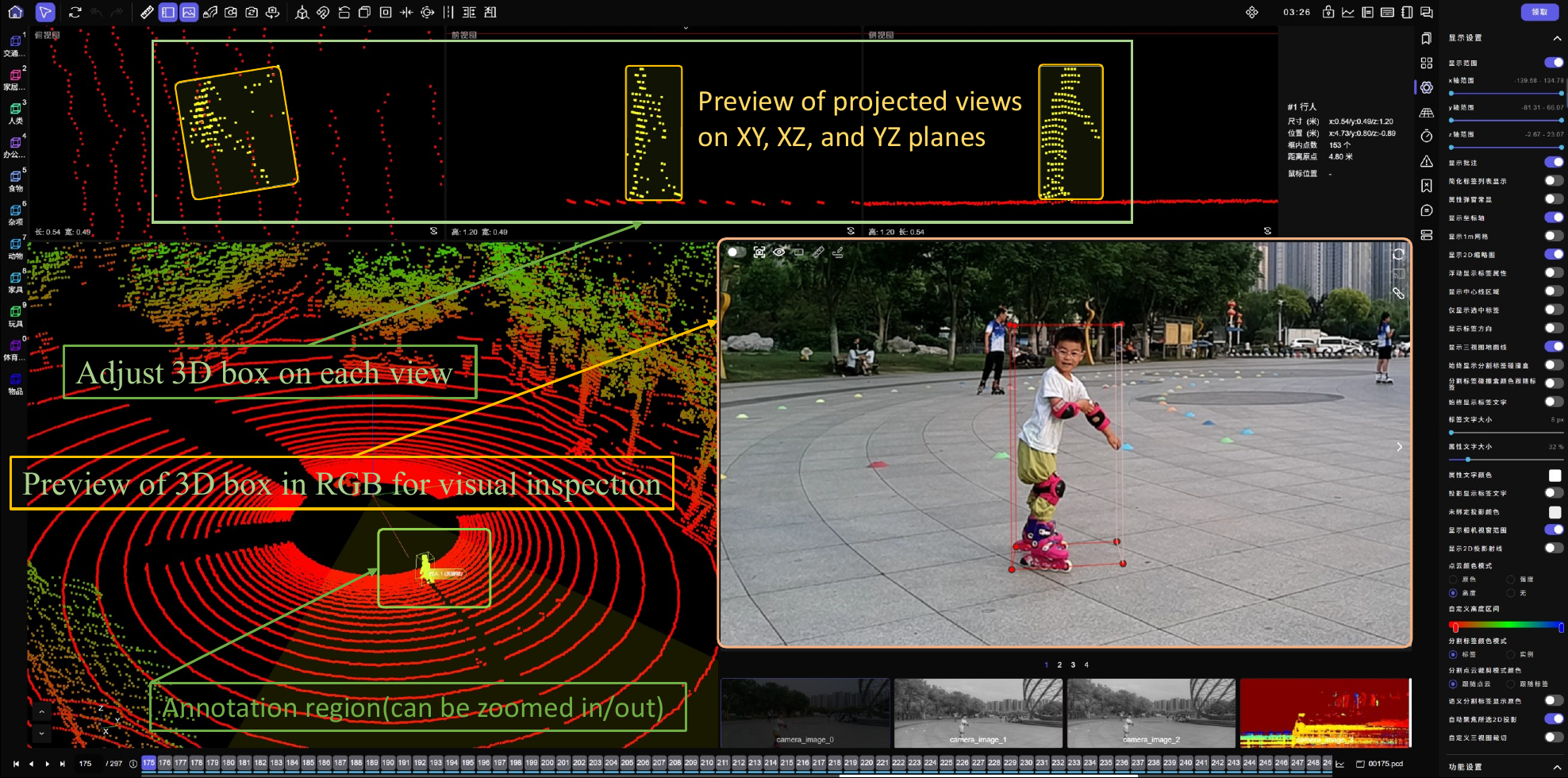}
    \caption{Annotation interface of our used annotation tool.}
    \label{fig:molardata}
\end{figure*}
The 3D annotation in this study was completed using a professional annotation platform provided by a certain company. The interface for its 3D bounding box annotation is shown in Fig.~\ref{fig:molardata}. In practice, we annotated each frame of the point cloud on a continuous trajectory segment basis: the annotator first sketched the approximate 3D bounding box of the target in a scalable annotation view, completing the initial bounding; then, switching to the XY, XZ, and YZ projection views, the corresponding 2D bounding boxes were fine-tuned to ensure the 3D bounding boxes closely matched the target shape in the three orthogonal directions. Simultaneously, the tool projected the obtained 3D bounding boxes onto an RGB image, providing an intuitive visual preview so that the annotator could re-verify and fine-tune the results from an image perspective.

\subsection*{S2.1 \; Annotation Cost and Workflow}
The annotation process involved 20 members in total, including 16 for initial labeling and subsequent revision, and 4 for quality checking and result review. Before formal annotation started, we first screened the raw collected videos and removed video sequences that did not meet the requirements, such as cases where the point cloud was present but the target was not within the frustum range, the target was unclear and no visible target could be observed, the sensor data was abnormal, or the content did not satisfy the annotation requirements, so as to ensure the usability of the data for subsequent annotation. On this basis, all annotators received unified training to ensure a consistent understanding of annotation standards, bounding box definition, and operation procedures. For each clip, we first annotated its first frame as an important reference for subsequent continuous-frame annotation, so as to improve annotation efficiency and maintain temporal consistency. The whole annotation process lasted about 8 months, and multiple rounds of checking and revision were used to further ensure the accuracy and reliability of the annotation results.


\section*{S3 \; More Statistics}
\begin{figure*}[!t]
    \centering
    \includegraphics[width=1.0\linewidth]{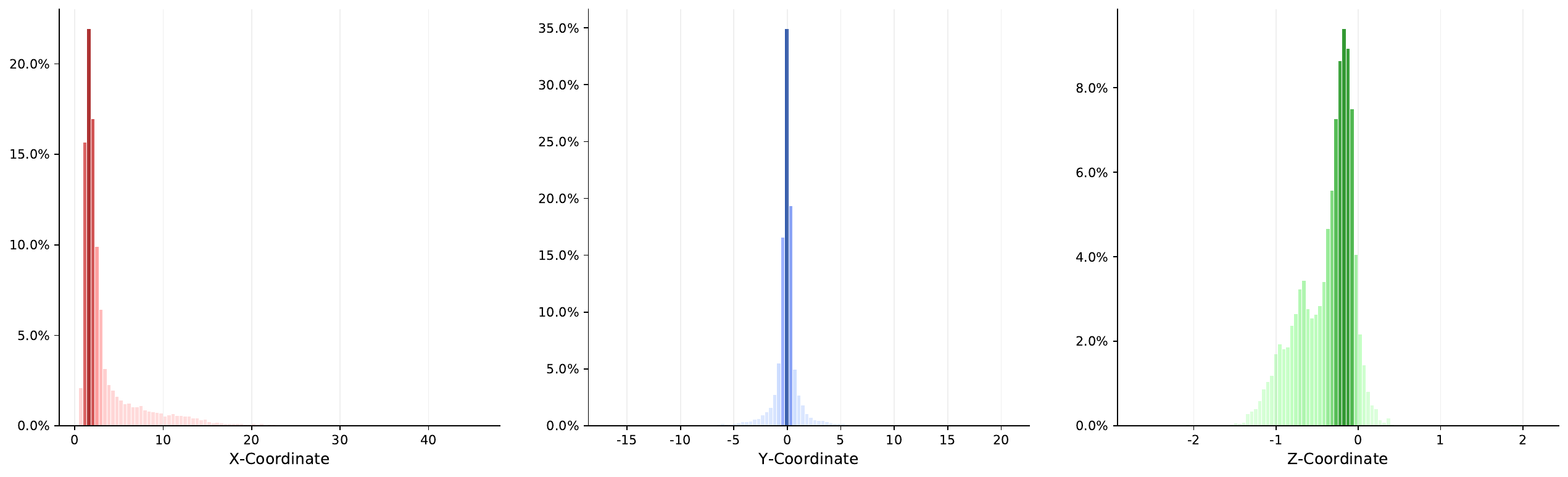}
    \vspace{-3mm}
    \caption{Statistics on 3DVQL. Distribution of target object center coordinates $(X, Y, Z)$ in 3D space.}
    \label{fig:statistic1}
\end{figure*}

\begin{figure*}[!t]
    \centering
    \includegraphics[width=1.0\linewidth]{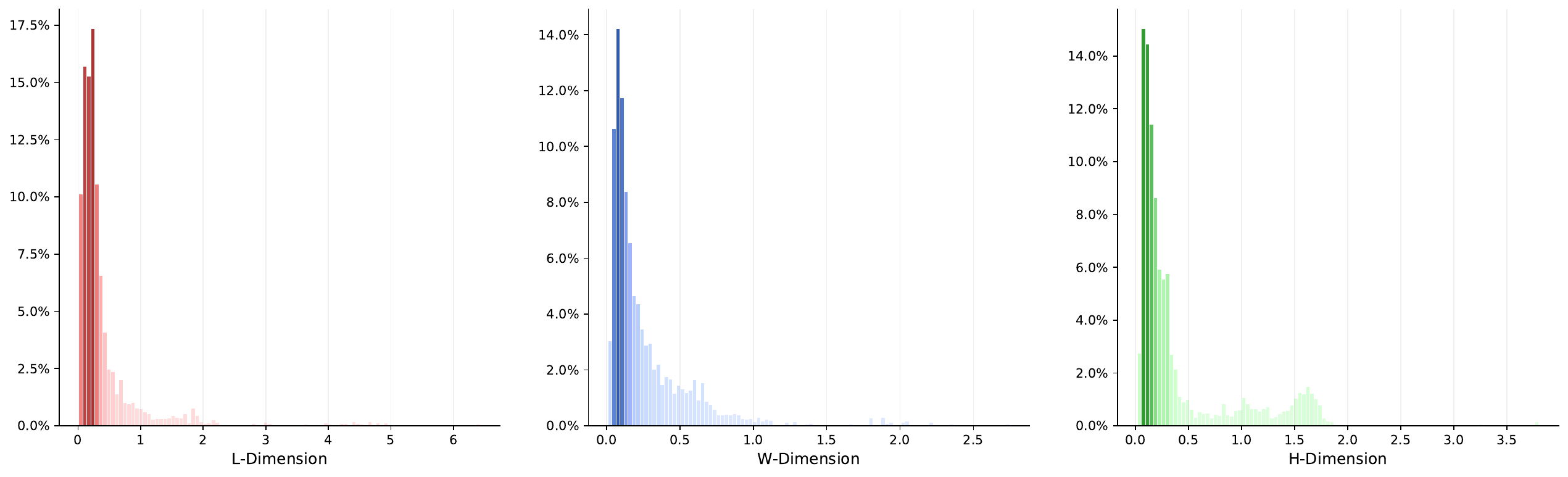}
    \vspace{-3mm}
    \caption{Statistics on 3DVQL. Distribution of target object sizes in length, width, and height $(L, W, H)$.}
    \label{fig:statistic2}
\end{figure*}

\begin{figure*}[!t]
    \centering
    \includegraphics[width=1.0\linewidth]{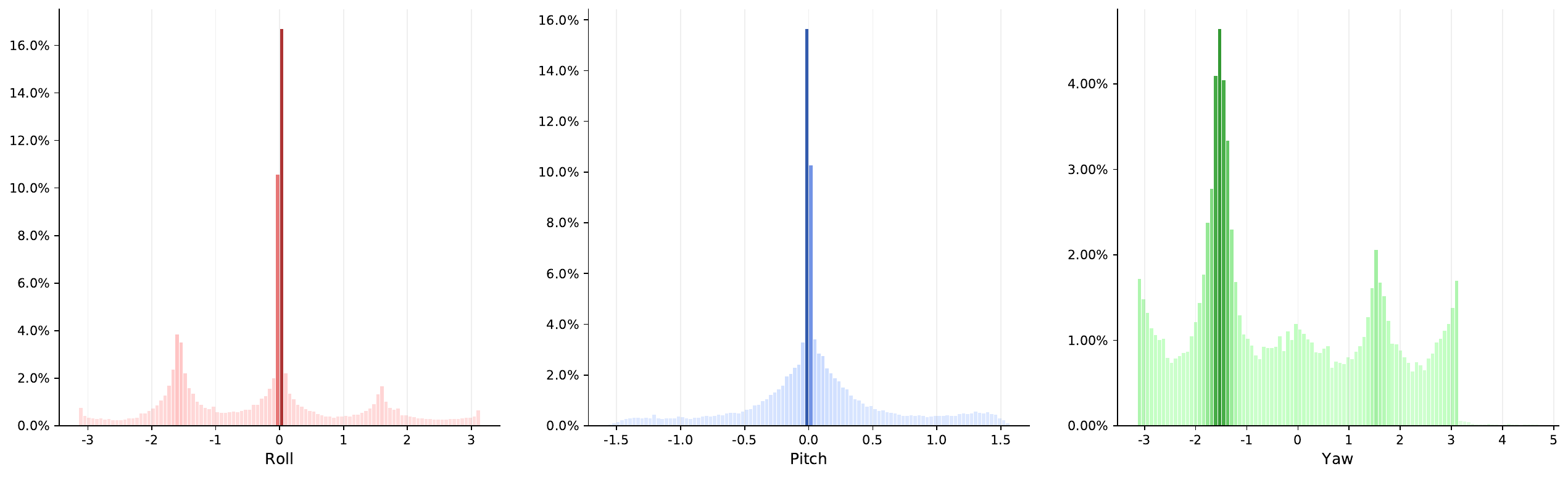}
    \vspace{-3mm}
    \caption{Spatial range statistics on 3DVQL: distribution of target object orientations in Roll, Pitch, and Yaw.}
    \label{fig:statistic3}
\end{figure*}

In this section, we further analyze the spatial statistics of 3DVQL. Specifically, Fig.~\ref{fig:statistic1} shows the distribution of the 3D center coordinates of all targets on the $X$, $Y$, and $Z$ axes. It can be seen that the targets are mainly concentrated in the region $X!\in[0,10]$\,m, $Y!\in[-2,2]$\,m, and $Z!\in[-1,1]$\,m. This is also the spatial range setting we adopted in our implementation: $\texttt{space\_range}=[[0.0,-2.0,-1.0], [10.0,2.0,1.0]]$. Simultaneously, we evenly divided this range into $16\times16\times16$ grid cells in the three directions ($\texttt{X\_center\_num}=\texttt{Y\_center\_num}=\texttt{Z\_center\_num}=16$) for subsequent 3D retrieval and localization. Fig.~\ref{fig:statistic2} shows the size distribution of the target's 3D bounding box along the three dimensions of length ($L$), width ($W$), and height ($H$). It can be seen that most targets are concentrated in a relatively small scale, but still cover a variety of sizes from everyday small objects to medium-sized facilities. Fig.~\ref{fig:statistic3} statistically analyzes the distribution of target poses in the three angles of roll, pitch, and yaw. Roll and pitch are approximately symmetrical around $0^\circ$, while yaw exhibits several dominant orientations. 

\begin{figure*}[!t]
    \centering
    \includegraphics[width=1.0\linewidth]{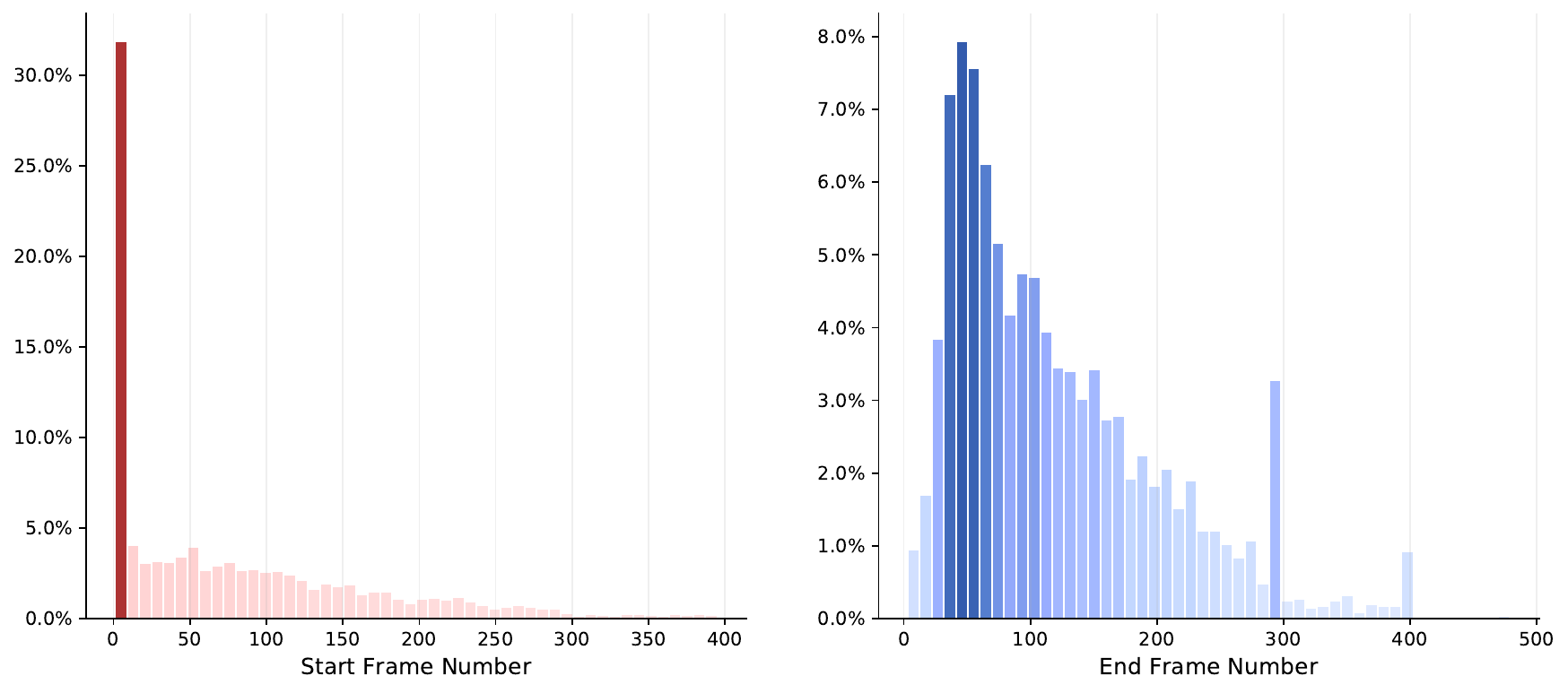}
    \vspace{-3mm}
    \caption{Statistics on 3DVQL: Distribution of the start and end frame indices of response tracklets.}
    \label{fig:statistic4}
\end{figure*}

As shown in Fig.~\ref{fig:statistic4}, we count the distributions of the indices of the starting frames and ending frames of the response tracklets. At the same time, we can clearly observe that the indices of the starting frames are mainly concentrated from frame 0 to frame 200, which indicates that most queries can obtain a response in the early stage of the video, although there are still some cases after frame 200, which means that higher temporal robustness is required for the query algorithms. The indices of the ending frames mostly fall between frame 40 and frame 400, and there is also a certain long tail, indicating that the lengths of the response tracklets in the dataset exhibit a clear long tail distribution.

These distributions also explain why we used a 9 DoF threshold of 0.05. We also hope that this statistical information will help readers gain a more comprehensive understanding of the characteristics of the 3DVQL dataset in terms of spatial range, object scale, and pose diversity.

\subsection*{S3.1 \; Video and Response Statistics}
We define the video length as the number of frames in which the target appears. Under 20 FPS, this length ranges from 40 to 390 frames, with an average of 86 frames. The statistics here reflect the duration of target visibility, rather than the full clip length including negative sample segments. In addition, each sequence contains 1 to 5 response segments, with an average of about 3 response segments per sequence. This shows that the target is usually visible intermittently rather than continuously within one interval, and this also makes cross-segment localization a key property of 3DVQL. The nature of 3DVQL is not continuous tracking in the traditional sense, but query-driven target localization in complex video scenes. Given a query, the model needs to complete target identification and localization across multiple discrete response segments, and handle the matching challenges caused by target disappearance, reappearance, and viewpoint changes.


\subsection*{S3.2 \; Short Sequences and Episodic Memory}
When the robot switches areas or the target leaves the field of view, it usually means the clip ends. This keeps the sequence length short and also provides enough viewpoint changes for learning negative samples. This segmentation also naturally covers cases where the target is temporarily absent, re-enters the field of view, or recovers from occlusion. Sequences with negative samples have an average length of 311 frames, which is about 15.5 seconds at 20 fps, and are suitable for 3D visual query localization with frequent occlusions and viewpoint changes. In future work, we will add 200 longer sequences for long-term 3D VQL research.




\section*{S4 \; Evaluation Metrics and 3D IoU}

\vspace{0.45em}
\noindent
\textbf{Evaluation Protocol.}
Inspired by the 2DVQL evaluation protocol in~\cite{grauman2022ego4d}, we define the following metrics for 3D visual query localization under the Top-1 retrieval setting. For each query, the model outputs a single 3D response track with an associated confidence score. Average Precision (AP) is computed by ranking predictions according to confidence and integrating the precision–recall curve in the standard way.

Let $\mathcal{Q}$ be the set of all queries and $|\mathcal{Q}|$ its cardinality. We use $\mathcal{T} = \{0.25, 0.50, 0.75, 0.95\}$ for temporal IoU (tIoU) thresholds and $\mathcal{S} = \{0.05, 0.25, 0.50, 0.75, 0.95\}$ for 3D spatio-temporal IoU ($\mathrm{stIoU}_{3D}$) thresholds.

\vspace{0.45em}
\noindent
\textbf{Temporal Average Precision (tAP).}
tAP measures how well the predicted temporal interval matches the ground-truth response interval. For a given tIoU threshold $\delta \in \mathcal{T}$, we compute the temporal AP, denoted as $\text{AP}^{t}(\delta)$. The final tAP is obtained by averaging over all tIoU thresholds:
\begin{equation}
    \setlength{\abovedisplayskip}{4pt} 
    \setlength{\belowdisplayskip}{4pt}
    \text{tAP} = \frac{1}{|\mathcal{T}|} \sum_{\delta \in \mathcal{T}} \text{AP}^{t}(\delta).
\end{equation}

\vspace{0.45em}
\noindent
\textbf{3D Spatio-Temporal Average Precision (3D-stAP).}
3D-stAP evaluates how well the 3D spatio-temporal tube of the prediction aligns with the ground-truth response track. For query $q \in \mathcal{Q}$, let $\mathcal{F}_q$ be the set of frames within its ground-truth response interval, and $\mathrm{IoU}_{3D}^{q,t}$ the 3D IoU between the predicted and ground-truth oriented 3D bounding boxes at frame $t \in \mathcal{F}_q$. Each 3D box is parameterized by 9 degrees of freedom,
\begin{equation}
    b = (x, y, z, l, w, h, \text{yaw}, \text{pitch}, \text{roll}),
\end{equation}
where $(x,y,z)$ denotes the 3D center, $(l,w,h)$ the box size, and (yaw, pitch, roll) the three rotation angles around the vertical, lateral, and longitudinal axes, respectively. We first aggregate frame-wise 3D IoU over time to obtain the 3D spatio-temporal IoU,
\begin{equation}
    \setlength{\abovedisplayskip}{4pt} 
    \setlength{\belowdisplayskip}{4pt}
    \mathrm{stIoU}_{3D}(q) = \frac{1}{|\mathcal{F}_q|} \sum_{t \in \mathcal{F}_q} \mathrm{IoU}_{3D}^{q,t}.
\end{equation}
For a given $\tau \in \mathcal{S}$, we then compute the AP at 3D spatio-temporal IoU threshold $\tau$, denoted as $\text{AP}^{3\text{D-st}}(\tau)$. The 3D-stAP is defined as
\begin{equation}
    \setlength{\abovedisplayskip}{4pt} 
    \setlength{\belowdisplayskip}{4pt}
    \text{3D-stAP} = \frac{1}{|\mathcal{S}|} \sum_{\tau \in \mathcal{S}} \text{AP}^{3\text{D-st}}(\tau).
\end{equation}
Given the increased difficulty of full 9-DoF localization (position, scale, yaw, pitch, and roll) in 3D space, we include a more permissive low threshold $\tau = 0.05$ in $\mathcal{S}$.

\vspace{0.45em}
\noindent
\textbf{Success (Succ).}
Succ measures whether the prediction achieves any effective overlap with the ground truth in 3D spatio-temporal space. A query $q$ is regarded as successful if its 3D spatio-temporal IoU satisfies $\mathrm{stIoU}_{3D}(q) \ge 0.05$. The overall success is the fraction of successful queries:
\begin{equation}
    \setlength{\abovedisplayskip}{4pt} 
    \setlength{\belowdisplayskip}{4pt}
    \text{Succ} = \frac{1}{|\mathcal{Q}|} \sum_{q \in \mathcal{Q}} \mathbb{I}\big( \mathrm{stIoU}_{3D}(q) \ge 0.05 \big),
\end{equation}
where $\mathbb{I}(\cdot)$ is the indicator function.

\vspace{0.45em}
\noindent
\textbf{Recovery\% (Rec\%).}
Rec\% focuses on how much of the ground-truth response track is recovered at the frame level. For each query $q$, let $\mathcal{F}_q$ be its ground-truth response frames as above. We count a frame $t \in \mathcal{F}_q$ as recovered if $\mathrm{IoU}_{3D}^{q,t} \ge 0.5$. Rec\% is defined as the percentage of such recovered frames among all frames on all response tracks:
\begin{equation}
    \setlength{\abovedisplayskip}{4pt} 
    \setlength{\belowdisplayskip}{4pt}
    \text{Rec\%} = 
    \frac{\displaystyle \sum_{q \in \mathcal{Q}} \sum_{t \in \mathcal{F}_q} \mathbb{I}\big( \mathrm{IoU}_{3D}^{q,t} \ge 0.5 \big)}
         {\displaystyle \sum_{q \in \mathcal{Q}} |\mathcal{F}_q|} 
    \times 100\%.
\end{equation}
This metric follows the robustness spirit of the VOT challenge~\cite{kristan2020eighth} by emphasizing stable recovery of the target over time.

\section*{S5 \; Details of Baselines}

\begin{figure*}[!t]
    \centering
    \includegraphics[width=0.95\linewidth]{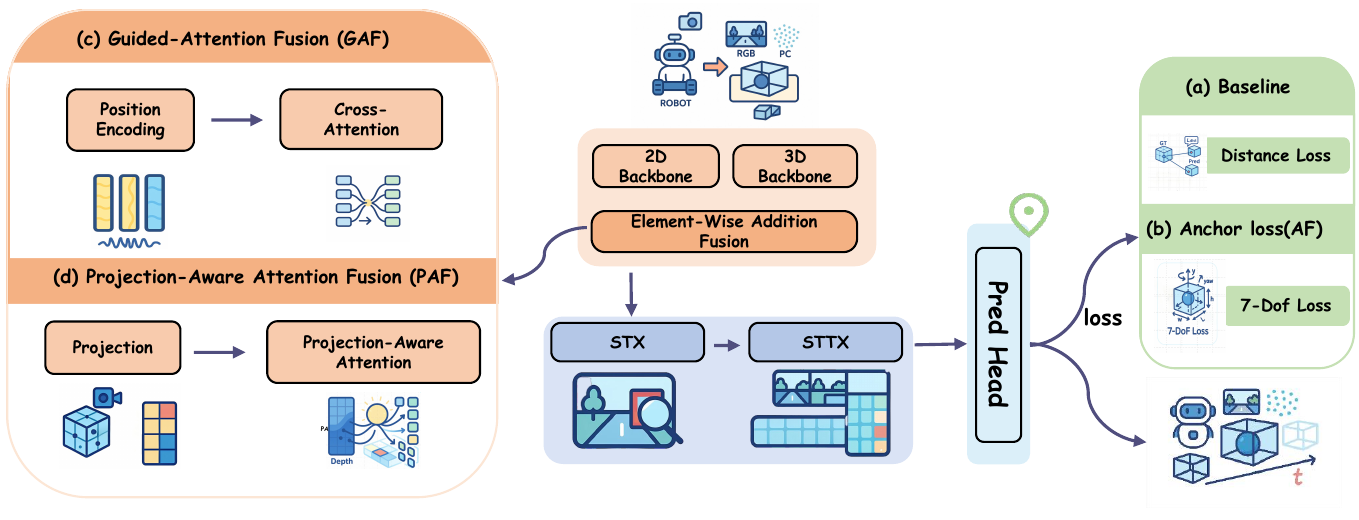}
    \caption{Architectural comparison of the baseline variants: (a) Baseline, (b) AnchorFusion-3DVQL (AF), (c) Guided-Attention Fusion (GAF), (d) Projection-Aware Attention Fusion (PAF).}
    \label{fig:baseline}
\end{figure*}

In this section, we present a detailed elaboration on the baseline variants introduced in the main text. Given that the overall architecture has been previously outlined, the following discussion focuses specifically on the implementation details of the modifications in each variant.

\paragraph{AnchorFusion-3DVQL (AF)} As illustrated in Fig.~\ref{fig:baseline}(b), the absence of a differentiable 9-DoF IoU operator impedes stable supervision across multiple candidate boxes. To address this, we adopt a 7-DoF Generalized IoU (GIoU) loss as the training objective, replacing the standard center-point regression loss. Specifically, the roll and pitch parameters are zeroed out during the GIoU calculation between ground truth and predicted boxes. The total loss function is formulated as:
\begin{equation}
\mathcal{L} = \lambda_{\text{c}} \mathcal{L}_{c} + \lambda_{\text{s}} \mathcal{L}_{s} + \lambda_{\text{r}} \mathcal{L}_{r} + \lambda_{\text{cls}} \mathcal{L}_{\text{cls}} + \lambda_{\text{GIoU}} \mathcal{L}_{\text{GIoU}}.
\label{eq:7dofloss}
\end{equation}

\paragraph{Guided-Attention Fusion-3DVQL (GAF)} As depicted in Fig.~\ref{fig:baseline}(c), this variant integrates a Guided-Attention Fusion module designed to explicitly inject depth cues via depth positional encodings and depth-aware blocks. In this setup, learnable parameters are augmented to the 3D features along the depth axis. Subsequently, the 2D and 3D features are fused through a depth-wise cross-attention mechanism.

\paragraph{Projection-Aware Attention Fusion-3DVQL (PAF)} As shown in Fig.~\ref{fig:baseline}(d), this variant incorporates a Projection-Aware Attention Fusion module. Here, the centers of 3D voxels are projected onto the 2D image plane via the camera matrix, after which corresponding image features are sampled using bilinear interpolation. These sampled 2D features are then fused with the 3D voxel features via a multi-head attention mechanism along the depth axis, where the 3D features serve as queries and the sampled 2D features serve as keys and values.

\subsection*{S5.1 \; Additional Modality Settings and Broader Baselines}
Although this paper mainly designs and evaluates baseline models under the RGB--PC setting, in order to further verify the role of the dataset, we supplement more modality settings, including RGB-only, PC-only, RGB-D, and PC-D. This makes it possible to more directly compare the effects of appearance information, geometric information, and their combinations in 3D visual query localization. In addition, we also include the point-tracking-style baseline TAPIP3D~\cite{zhang2025tapip3d}, as well as the two-stage baseline TWO-Stage built upon CenterPoint~\cite{yin2021center} and MBPTracker~\cite{xu2023mbptrack}, for supplementary comparison with single-stage fusion methods. The results are shown in Tab.~\ref{tab:more_baselines}. The first four rows are modality variants based on LaF, and the last two rows are broader comparison baselines. It can be seen that, in embodied environments, using only RGB or Depth often leads to insufficient model learning, while settings with point cloud information usually perform better. This further shows the reasonableness of the RGB--PC fusion setting. At the same time, the results of TAPIP3D and TWO-Stage also show that 3DVQL is challenging for methods of different paradigms.

\begin{table}[t]
    \centering
    \small
    \caption{Additional modality settings and broader baselines on 3DVQL. The first four rows correspond to modality variants, while the last two rows report broader baselines beyond the single-stage fusion framework.}
    \vspace{-3mm}
    \resizebox{0.48\textwidth}{!}{
    \begin{tabular}{lcccccc}
        \toprule
        \textbf{Method} & \textbf{tAP} & $\mathbf{tAP}_{0.25}$ & \textbf{stAP} & $\mathbf{stAP}_{0.05}$ & \textbf{Rec.\%} & \textbf{Succ.} \\
        \midrule
        LaF (RGB-only) & 0.007 & 0.028 & 0.003 & 0.015 & 0.020 & 12.302 \\
        LaF (PC-only)  & 0.030 & 0.505 & 0.012 & 0.062 & 0.095 & 26.481 \\
        LaF (RGB-D)    & 0.135 & 0.335 & 0.006 & 0.035 & 0.055 & 16.516 \\
        LaF (PC-D)     & 0.225 & 0.545 & 0.015 & 0.075 & 0.115 & 30.632 \\
        TAPIP3D        & 0.143 & 0.485 & 0.007 & 0.041 & 0.063 & 18.824 \\
        TWO-Stage      & 0.162 & 0.411 & 0.009 & 0.045 & 0.075 & 21.431 \\
        \bottomrule
    \end{tabular}}
    \label{tab:more_baselines}
    \vspace{-3mm}
\end{table}

\section*{S6 \; Qualitative Results} In this section, we present the visualization results of several proposed baseline methods (PAF, GAF, AF) and our method LaF in more scenarios using 3D VQL in Figure ~\ref{fig:supp_vis}. As can be seen from Fig.~\ref{fig:supp_vis}, these baseline methods often struggle to consistently and accurately locate targets in 3D space in complex scenes with frequent occlusion and similar interference. In contrast, LaF can more stably complete spatial query and localization in such situations, demonstrating better robustness and adaptability.

\section*{S7 \; Maintenance and Responsible Usage of 3DVQL for Research}

\textbf{Maintenance.} We will host 3DVQL on the widely used GitHub platform (where all dataset download links and our models will be publicly released). This allows us to promptly view feedback from the community and answer all user questions. It also enables us to maintain and update our benchmarks as needed by researchers, thus continuously improving them. Simultaneously, the authors will make every effort to collect and organize evaluation results of various algorithms on 3DVQL, forming a relatively complete and dynamically updated comparative analysis. Our ultimate goal is for 3DVQL to gradually develop into a long-term, sustainable public platform for multimodal 3D visual query localization research.



\vspace{0.3em}
\noindent
\textbf{Responsible Usage of 3DVQL.} 3DVQL aims to promote research and application related to three-dimensional visual query localization. It is development and use are limited to \textbf{\emph{research purpose only}}.

{
    \small
    \bibliographystyle{ieeenat_fullname}
    \bibliography{main}

@String(CVPR= {IEEE Conf. Comput. Vis. Pattern Recog.})

@String(ICLR = {Int. Conf. Learn. Represent.})

@String(CVPR  = {CVPR})

@String(ICLR  = {ICLR})

@inproceedings{li2024mvbench,
  title={Mvbench: A comprehensive multi-modal video understanding benchmark},
  author={Li, Kunchang and Wang, Yali and He, Yinan and Li, Yizhuo and Wang, Yi and Liu, Yi and Wang, Zun and Xu, Jilan and Chen, Guo and Luo, Ping and others},
  booktitle={Proceedings of the IEEE/CVF Conference on Computer Vision and Pattern Recognition},
  pages={22195--22206},
  year={2024}
}

@inproceedings{li2023uniformerv2,
  title={Uniformerv2: Unlocking the potential of image vits for video understanding},
  author={Li, Kunchang and Wang, Yali and He, Yinan and Li, Yizhuo and Wang, Yi and Wang, Limin and Qiao, Yu},
  booktitle={Proceedings of the IEEE/CVF International Conference on Computer Vision},
  pages={1632--1643},
  year={2023}
}

@inproceedings{ma2025position,
  title={Position: Prospective of autonomous driving-multimodal llms world models embodied intelligence ai alignment and mamba},
  author={Ma, Yunsheng and Ye, Wenqian and Cui, Can and Zhang, Haiming and Xing, Shuo and Ke, Fucai and Wang, Jinhong and Miao, Chenglin and Chen, Jintai and Rezatofighi, Hamid and others},
  booktitle={Proceedings of the Winter Conference on Applications of Computer Vision},
  pages={1010--1026},
  year={2025}
}

@inproceedings{wang2024embodiedscan,
  title={Embodiedscan: A holistic multi-modal 3d perception suite towards embodied ai},
  author={Wang, Tai and Mao, Xiaohan and Zhu, Chenming and Xu, Runsen and Lyu, Ruiyuan and Li, Peisen and Chen, Xiao and Zhang, Wenwei and Chen, Kai and Xue, Tianfan and others},
  booktitle={Proceedings of the IEEE/CVF Conference on Computer Vision and Pattern Recognition},
  pages={19757--19767},
  year={2024}
}

@article{dosovitskiy2020vit,
  title={An Image is Worth 16x16 Words: Transformers for Image Recognition at Scale},
  author={Dosovitskiy, Alexey and Beyer, Lucas and Kolesnikov, Alexander and Weissenborn, Dirk and Zhai, Xiaohua and Unterthiner, Thomas and  Dehghani, Mostafa and Minderer, Matthias and Heigold, Georg and Gelly, Sylvain and Uszkoreit, Jakob and Houlsby, Neil},
  journal={ICLR},
  year={2021}
}

@inproceedings{grauman2022ego4d,
  title={Ego4d: Around the world in 3,000 hours of egocentric video},
  author={Grauman, Kristen and Westbury, Andrew and Byrne, Eugene and Chavis, Zachary and Furnari, Antonino and Girdhar, Rohit and Hamburger, Jackson and Jiang, Hao and Liu, Miao and Liu, Xingyu and others},
  booktitle={Proceedings of the IEEE/CVF conference on computer vision and pattern recognition},
  pages={18995--19012},
  year={2022}
}

@article{fan2025prvql,
  title={Prvql: Progressive knowledge-guided refinement for robust egocentric visual query localization},
  author={Fan, Bing and Feng, Yunhe and Tian, Yapeng and Liang, James Chenhao and Lin, Yuewei and Huang, Yan and Fan, Heng},
  journal={arXiv preprint arXiv:2502.07707},
  year={2025}
}

@article{jiang2023single,
  title={Single-stage visual query localization in egocentric videos},
  author={Jiang, Hanwen and Ramakrishnan, Santhosh Kumar and Grauman, Kristen},
  journal={Advances in Neural Information Processing Systems},
  volume={36},
  pages={24143--24157},
  year={2023}
}

@article{xu2022negative,
  title={Negative frames matter in egocentric visual query 2d localization},
  author={Xu, Mengmeng and Fu, Cheng-Yang and Li, Yanghao and Ghanem, Bernard and Perez-Rua, Juan-Manuel and Xiang, Tao},
  journal={arXiv preprint arXiv:2208.01949},
  year={2022}
}

@inproceedings{xu2023my,
  title={Where is my wallet? modeling object proposal sets for egocentric visual query localization},
  author={Xu, Mengmeng and Li, Yanghao and Fu, Cheng-Yang and Ghanem, Bernard and Xiang, Tao and P{\'e}rez-R{\'u}a, Juan-Manuel},
  booktitle={Proceedings of the IEEE/CVF Conference on Computer Vision and Pattern Recognition},
  pages={2593--2603},
  year={2023}
}

@inproceedings{sun2020scalability,
  title={Scalability in perception for autonomous driving: Waymo open dataset},
  author={Sun, Pei and Kretzschmar, Henrik and Dotiwalla, Xerxes and Chouard, Aurelien and Patnaik, Vijaysai and Tsui, Paul and Guo, James and Zhou, Yin and Chai, Yuning and Caine, Benjamin and others},
  booktitle={Proceedings of the IEEE/CVF conference on computer vision and pattern recognition},
  pages={2446--2454},
  year={2020}
}

@inproceedings{geiger2012we,
  title={Are we ready for autonomous driving? the kitti vision benchmark suite},
  author={Geiger, Andreas and Lenz, Philip and Urtasun, Raquel},
  booktitle={CVPR},
  year={2012}
}

@inproceedings{caesar2020nuscenes,
  title={nuscenes: A multimodal dataset for autonomous driving},
  author={Caesar, Holger and Bankiti, Varun and Lang, Alex H and Vora, Sourabh and Liong, Venice Erin and Xu, Qiang and Krishnan, Anush and Pan, Yu and Baldan, Giancarlo and Beijbom, Oscar},
  booktitle={CVPR},
  year={2020}
}

@inproceedings{khosla2025relocate,
  title={Relocate: A simple training-free baseline for visual query localization using region-based representations},
  author={Khosla, Savya and Schwing, Alexander and Hoiem, Derek and others},
  booktitle={Proceedings of the Computer Vision and Pattern Recognition Conference},
  pages={3697--3706},
  year={2025}
}

@inproceedings{yang2022towards,
  title={Towards generic 3d tracking in RGBD videos: Benchmark and baseline},
  author={Yang, Jinyu and Zhang, Zhongqun and Li, Zhe and Chang, Hyung Jin and Leonardis, Ale{\v{s}} and Zheng, Feng},
  booktitle={European Conference on Computer Vision},
  pages={112--128},
  year={2022},
  organization={Springer}
}

@inproceedings{song2015sun,
  title={Sun rgb-d: A rgb-d scene understanding benchmark suite},
  author={Song, Shuran and Lichtenberg, Samuel P and Xiao, Jianxiong},
  booktitle={Proceedings of the IEEE conference on computer vision and pattern recognition},
  pages={567--576},
  year={2015}
}

@inproceedings{dai2017scannet,
  title={Scannet: Richly-annotated 3d reconstructions of indoor scenes},
  author={Dai, Angela and Chang, Angel X and Savva, Manolis and Halber, Maciej and Funkhouser, Thomas and Nie{\ss}ner, Matthias},
  booktitle={Proceedings of the IEEE conference on computer vision and pattern recognition},
  pages={5828--5839},
  year={2017}
}

@inproceedings{zhou2018voxelnet,
  title={Voxelnet: End-to-end learning for point cloud based 3d object detection},
  author={Zhou, Yin and Tuzel, Oncel},
  booktitle={Proceedings of the IEEE conference on computer vision and pattern recognition},
  pages={4490--4499},
  year={2018}
}

@inproceedings{shi2019pointrcnn,
  title={Pointrcnn: 3d object proposal generation and detection from point cloud},
  author={Shi, Shaoshuai and Wang, Xiaogang and Li, Hongsheng},
  booktitle={Proceedings of the IEEE/CVF conference on computer vision and pattern recognition},
  pages={770--779},
  year={2019}
}

@inproceedings{qi2017pointnet,
  title={Pointnet: Deep learning on point sets for 3d classification and segmentation},
  author={Qi, Charles R and Su, Hao and Mo, Kaichun and Guibas, Leonidas J},
  booktitle={Proceedings of the IEEE conference on computer vision and pattern recognition},
  pages={652--660},
  year={2017}
}

@inproceedings{yin2021center,
  title={Center-based 3d object detection and tracking},
  author={Yin, Tianwei and Zhou, Xingyi and Krahenbuhl, Philipp},
  booktitle={Proceedings of the IEEE/CVF conference on computer vision and pattern recognition},
  pages={11784--11793},
  year={2021}
}

@inproceedings{chen2022focal,
  title={Focal sparse convolutional networks for 3d object detection},
  author={Chen, Yukang and Li, Yanwei and Zhang, Xiangyu and Sun, Jian and Jia, Jiaya},
  booktitle={Proceedings of the IEEE/CVF conference on computer vision and pattern recognition},
  pages={5428--5437},
  year={2022}
}

@inproceedings{mousavian20173d,
  title={3d bounding box estimation using deep learning and geometry},
  author={Mousavian, Arsalan and Anguelov, Dragomir and Flynn, John and Kosecka, Jana},
  booktitle={Proceedings of the IEEE conference on Computer Vision and Pattern Recognition},
  pages={7074--7082},
  year={2017}
}

@inproceedings{chen2016monocular,
  title={Monocular 3d object detection for autonomous driving},
  author={Chen, Xiaozhi and Kundu, Kaustav and Zhang, Ziyu and Ma, Huimin and Fidler, Sanja and Urtasun, Raquel},
  booktitle={Proceedings of the IEEE conference on computer vision and pattern recognition},
  pages={2147--2156},
  year={2016}
}

@inproceedings{ku2019monocular,
  title={Monocular 3d object detection leveraging accurate proposals and shape reconstruction},
  author={Ku, Jason and Pon, Alex D and Waslander, Steven L},
  booktitle={Proceedings of the IEEE/CVF conference on computer vision and pattern recognition},
  pages={11867--11876},
  year={2019}
}

@inproceedings{vora2020pointpainting,
  title={Pointpainting: Sequential fusion for 3d object detection},
  author={Vora, Sourabh and Lang, Alex H and Helou, Bassam and Beijbom, Oscar},
  booktitle={Proceedings of the IEEE/CVF conference on computer vision and pattern recognition},
  pages={4604--4612},
  year={2020}
}

@inproceedings{shi2019point,
  title={Point-voxel feature set abstraction for 3d object detection. 2020 IEEE},
  author={Shi, Shaoshuai and Guo, Chaoxu and Jiang, Li and Wang, Zhe and Shi, Jianping and Wang, Xiaogang and Pv-rcnn, Hongsheng Li},
  booktitle={CVF Conference on Computer Vision and Pattern Recognition (CVPR)},
  pages={10526--10535},
  year={2019}
}

@inproceedings{xu2023mbptrack,
  title={Mbptrack: Improving 3d point cloud tracking with memory networks and box priors},
  author={Xu, Tian-Xing and Guo, Yuan-Chen and Lai, Yu-Kun and Zhang, Song-Hai},
  booktitle={Proceedings of the IEEE/CVF International Conference on Computer Vision},
  pages={9911--9920},
  year={2023}
}

@inproceedings{caba2015activitynet,
  title={Activitynet: A large-scale video benchmark for human activity understanding},
  author={Caba Heilbron, Fabian and Escorcia, Victor and Ghanem, Bernard and Carlos Niebles, Juan},
  booktitle={Proceedings of the ieee conference on computer vision and pattern recognition},
  pages={961--970},
  year={2015}
}

@inproceedings{kristan2020eighth,
  title={The eighth visual object tracking VOT2020 challenge results},
  author={Kristan, Matej and Leonardis, Ale{\v{s}} and Matas, Ji{\v{r}}{\'\i} and Felsberg, Michael and Pflugfelder, Roman and K{\"a}m{\"a}r{\"a}inen, Joni-Kristian and Danelljan, Martin and Zajc, Luka {\v{C}}ehovin and Luke{\v{z}}i{\v{c}}, Alan and Drbohlav, Ondrej and others},
  booktitle={European conference on computer vision},
  pages={547--601},
  year={2020},
  organization={Springer}
}

@inproceedings{jiao2025gsot3d,
  title={GSOT3D: Towards Generic 3D Single Object Tracking in the Wild},
  author={Jiao, Yifan and Li, Yunhao and Ding, Junhua and Yang, Qing and Fu, Song and Fan, Heng and Zhang, Libo},
  booktitle={Proceedings of the IEEE/CVF International Conference on Computer Vision},
  pages={5469--5478},
  year={2025}
}

@article{dhall2017lidar,
  title={LiDAR-camera calibration using 3D-3D point correspondences},
  author={Dhall, Ankit and Chelani, Kunal and Radhakrishnan, Vishnu and Krishna, K Madhava},
  journal={arXiv},
  year={2017}
}

@article{oquab2023dinov2,
  title={Dinov2: Learning robust visual features without supervision},
  author={Oquab, Maxime and Darcet, Timoth{\'e}e and Moutakanni, Th{\'e}o and Vo, Huy and Szafraniec, Marc and Khalidov, Vasil and Fernandez, Pierre and Haziza, Daniel and Massa, Francisco and El-Nouby, Alaaeldin and others},
  journal={arXiv preprint arXiv:2304.07193},
  year={2023}
}

@article{zhang2025tapip3d,
  title={Tapip3d: Tracking any point in persistent 3d geometry},
  author={Zhang, Bowei and Ke, Lei and Harley, Adam W and Fragkiadaki, Katerina},
  journal={arXiv preprint arXiv:2504.14717},
  year={2025}
}
}


\end{document}